\documentclass[journal]{IEEEtai}

\usepackage{cite}

\ifCLASSINFOpdf

\RequirePackage[final]{graphicx}
\else 
\fi
\usepackage{pdflscape}
\usepackage{rotating}
\usepackage{epstopdf}
\usepackage{footnote}
\usepackage{verbatim} 
\usepackage{stfloats}
\usepackage{url}
\usepackage{float}
\usepackage{graphicx,algorithm,algorithmic,amsmath}
\usepackage{subfigure} 
\usepackage{cite}
\usepackage{multirow}
\usepackage{tabularx, booktabs}
\usepackage{multirow}
\usepackage{hhline}
\usepackage{units}
\usepackage{footnote}
\usepackage{threeparttable}
\usepackage{mathtools}
\usepackage{nicefrac}
\usepackage{amssymb}
\newcommand{\rpm}{\sbox0{$1$}\sbox2{$\scriptstyle\pm$}
	\raise\dimexpr(\ht0-\ht2)/2\relax\box2 }
\DeclareMathOperator*{\argminA}{argmin}
\allowdisplaybreaks
\begin{document}

\title{On fine-tuning of  Autoencoders  \\for Fuzzy rule classifiers}

\author{Rahul~Kumar~Sevakula,~\IEEEmembership{Member,~IEEE}
        ,~Nishchal~Kumar~Verma,~\IEEEmembership{Senior Member,~IEEE}
		and~Hisao~Ishibuchi,~\IEEEmembership{Fellow,~IEEE}

\thanks{E-mail: rsevakula@mgh.harvard.edu, nishchal@iitk.ac.in and hisaoi@cs.osakafu-u.ac.jp}
}

\maketitle

\begin{abstract}
Recent discoveries in Deep Neural Networks are allowing researchers to tackle some very complex problems such as image classification and audio classification, with improved theoretical and empirical justifications. This paper presents a novel scheme to incorporate the use of autoencoders in Fuzzy rule classifiers (FRC). Autoencoders when stacked can learn the complex non-linear relationships amongst data, and the proposed framework built towards FRC can allow users to input expert knowledge to the system. This paper further introduces four novel fine-tuning strategies for autoencoders to improve the FRC's classification and rule reduction performance. The proposed framework has been tested across five real-world benchmark datasets. Elaborate comparisons with over 15 previous studies, and across 10-fold cross validation performance, suggest that the proposed methods are capable of building FRCs which can provide state of the art accuracies.
\end{abstract}

\begin{IEEEImpStatement}
Many real-world classification problems necessitate that the artificial intelligence methods not only leverage the training data, but also be able to leverage any prior expert knowledge that is available, and/or they be explainable. FRCs are known to fulfill this need to a reasonable degree. This paper presents a well performing introductory framework that cohabits FRC and autoncoders (used in deep learning frameworks) together to leverage the advantages of both. Supporting the framework, a method to incorporate prior expert knowledge, and multiple finetuning strategies to further enhance the classification performance and rule-reduction performance, have been additionally introduced. The proposed framework provides very promising results on sample datasets; therefore this paper can be a catalyst to further the research in a direction where FRCs and deep learning intersect.

\end{IEEEImpStatement}


\section{Introduction}

Fuzzy rule based classifiers (FRCs) have been popular in real-world applications. Their popularity can be attributed to: 1) providing good classification performance, 2) providing facility to leverage existing expert knowledge and/or 3) providing a degree of explainability \cite{smc_howgoodarefrc}. FRCs were first introduced and popularized as a set of fuzzy If-Then rules for classification \cite{cordon,frc_ishibuchi1992,frc_ishibuchi1995}. 
The initial works on FRCs \cite{cordon} extended the Wang and Mendel's (WM) framework \cite{modelling_wm} to fuzzy classification rules, and made major contributions on how to weigh and aggregate the fuzzy rules.

FRCs in general suffer from a major limitation that the size of the rule space increases exponentially with the increase in the data dimensionality. 
Consider a dataset having $n$ features. When each axis of the $n$-dimensional pattern space is partitioned into $p$ fuzzy subsets, the total number of possible fuzzy rules is $p^n$.
Having such a large rule space not only increases the likelihood of generating more rules, but also leads to generalization issues \cite{tnnls_mvp} when data is insufficient. An FRC characterized by too many rules, is also susceptible to overfitting. These concerns are serious, and therefore numerous strategies have been developed over the years to address them.

Rule generation strategies for FRCs have typically been a multi-objective problem of finding optimal rule set(s), with maximization of classification accuracy and minimization of the number of rules \cite{fs_chen} as important objectives. Keeping this in mind, FRC frameworks are tuned in a variety of ways before use. A very common approach is to have measures/methods that judge the importance of individual rules and then select the optimal rule set \cite{frc_ishibuchi1995,reduction_jin,reduction_taniguchi,reduction_luo,reduction_sgerd,frc_ishibuchi2004,smc_perf_evaluation,ga_review}. These include the use of genetic algorithms for rule selection \cite{frc_ishibuchi1995,reduction_sgerd,frc_ishibuchi2004}, use of a fuzzy similarity measure to judge the importance of rules \cite{reduction_jin} and methods which use properties of the rule generation framework to identify and remove the redundant rules \cite{reduction_taniguchi,reduction_luo}. The second approach is to have measures/methods to judge the importance of features and select only the useful features \cite{fs_chen,fs_muni_genetic,smc_featureselection}. A third approach for reducing the number of rules and length of each rule, is to apply orthogonal transforms namely Singular Value Decomposition (SVD) on the rule consequent matrix \cite{reduction_yam,reduction_setnes,reduction_tao}. In the process of rule reduction, it is common for researchers to use interpolation methods to account for the lost rules \cite{reduction_koczy,reduction_tao}. A fourth approach for having optimal rule sets, is to have methods that optimize/tune the membership functions and add weights/confidence values to individual rules \cite{frc_ishibuchi2001,frc_ishibuchi2005}.

\begin{figure*}
	\centering
	\includegraphics[scale=0.37]{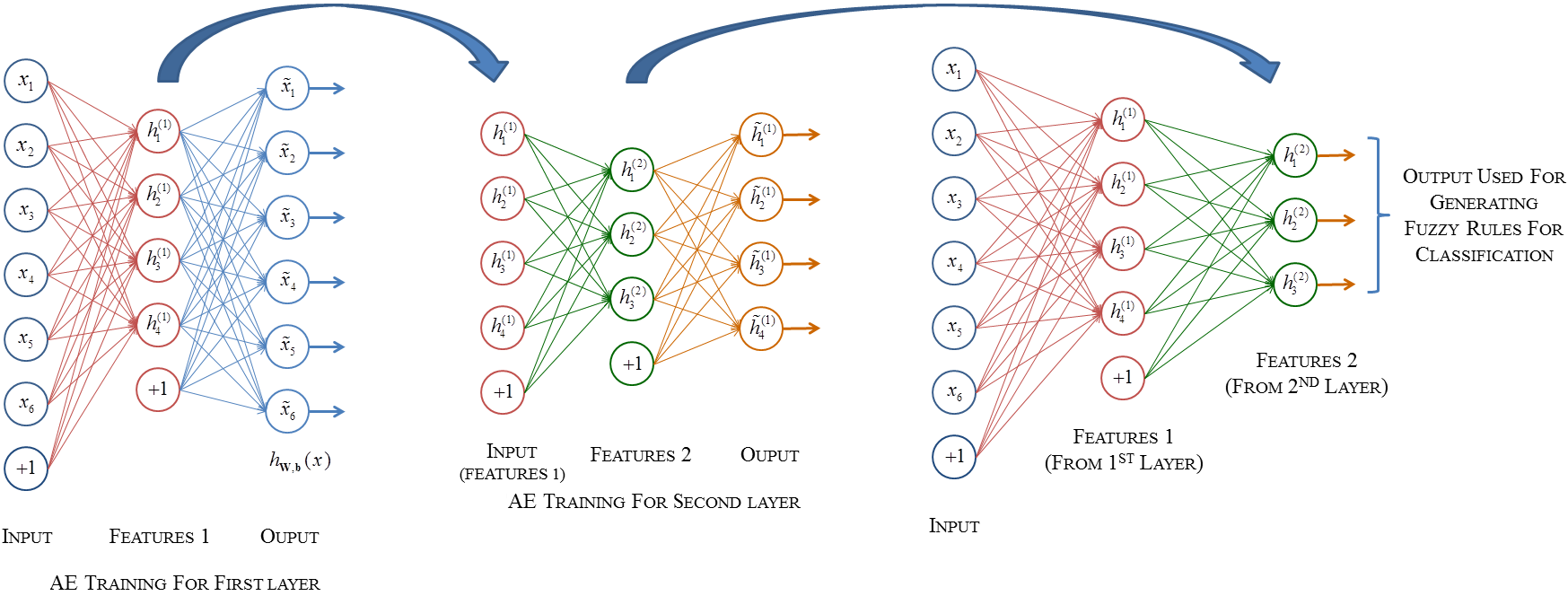}
	\vspace{-0.6em}
	\caption{Training of sparse autoencoders layer by layer for use in Stacked Sparse Autoencoder based FRC}
	\label{sae_sample}  
	\vspace{-1em}
\end{figure*}

Neuro-fuzzy inference systems form another paradigm in FRCs. Artificial Neural Networks (ANNs) carry major advantages, namely 1) flexibility to model complex problems and 2) ability to quickly learn the weights in the network using back-propagation based algorithms. These advantages make them an ideal tool for use in Fuzzy systems. Some neuro-fuzzy frameworks for FRC are SuPFuNIS \cite{neuro_SuPFuNIS}, FuGeNeSys \cite{neuro_fugenesys}, CNFS \cite{neuro_ctlin}, and a connectionist framework to select features while learning FRC \cite{neuro_debrup,neuro_debrup_connectionist}. The above frameworks make different modifications in the architecture of feedforward neural networks (FFNs), such that fuzzy logic is incorporated in the system, and its related parameters are learned through the back-propagation algorithm. All of them manage to give good classification accuracies with a compact rule base. However, there are following concerns with these approaches: 1) FFNs in the absence of regularization techniques, are susceptible to overfitting; therefore, appropriate attention is necessary while using similar architectures, 2) random initialization of parameters for FFNs can lead to sub-optimal solutions, especially when the FFN has more than one hidden layer \cite{dl_bengio2009}; therefore for complex data needing deeper FFNs, such architectures are inappropriate for use, 3) approaches given by \cite{neuro_debrup} and \cite{neuro_debrup_connectionist} are computationally too expensive and hence are not suited for larger datasets; this has been acknowledged by the authors themselves in the paper, 
and 4) there is no planned attempt to generate/learn improved feature representations such that classification by the FRC on non-separable complex data becomes more viable. 
The last concern is not only applicable to the neuro-fuzzy FRCs, but is applicable to most existing FRCs [2]-[4], [7]-[27]. 
This paper presents a novel neuro-fuzzy architecture for FRC, that specifically aims at resolving the last concern, and to a certain extent resolves the first and second concern too. Recently, \cite{deepCascadeFRC,deepTSK,deepFrcICU} proposed frameworks to develop deep rule based FRCs, however none of them use autoencoders, and none of them focus on obtaining an improved feature representation.

Existing FRC architectures perform operations directly on the input feature representation itself. Considering that deep learning has revolutionized the way of how features representations can be modeled/learned from data to facilitate classification of complex data \cite{dl_honglak_lee,dl_hinton2006,dl_hinton2012,dl_fuzzy_rbm,dl_railway}, 
this paper introduces a novel paradigm for generating compact and useful fuzzy rules with help of Autoencoders (AEs). An expression/function is called compact when it has fewer computational elements or fewer degrees of freedom for tuning \cite{dl_bengio2009}. AEs are used here to transform the input features to a compact feature space \cite{dl_bengio2009} that facilitates the FRC to have improved classification performance with few rules relatively. 
The AE based FRC has three components namely, 1) data preprocessing using fuzzy membership functions, 2) unsupervised greedy layerwise learning of AE, and 3) fine-tuning of the network weights and biases with methods introduced in this paper. Our earlier papers \cite{sevakula_sae1,sevakula_sae2} individually discussed the advantages of using Stacked Autoencoders for FRC \cite{sevakula_sae1}, and preprocessing the data before training an autoencoder based FRC \cite{sevakula_sae2}. This paper introduces an improvised end-to-end FRC framework with AE. The paper further contributes by introducing novel fine-tuning techniques which allow the proposed FRC framework to achieve significantly better classification performance. A strategy to incorporate expert knowledge into the system has also been introduced. For validating the proposed approach, the paper in experimentation section uses stacked AEs with denoising and sparsity abilities/constraints; such AEs are popularly known as Stacked Denoising Sparse Autoencoders (SDSAEs) \cite{dl_ng_lecture_notes,dl_denoising_ae}. The SDSAE based FRC was tested on five real-world benchmark datasets, and the same when measured through 10-fold cross-validation, was found to give better performance than 15 past FRC studies.

The paper proceeds as follows. Section II discusses our earlier attempts on developing autoencoder based FRC, and the limitations each attempt posed. The section then introduces the proposed AE based FRC framework, and discusses the procedure to introduce expert knowledge into the framework. Section III introduces novel strategies for fine-tuning the network's weights and biases. Section IV discusses the experiments performed on benchmark datasets. Section V concludes the paper.

\section{Autoencoder based FRC} 

In all pattern classification problems, the features play a significant role; if they provide appropriate distinction between data of different classes, a linear classifier would suffice to give good performance. Traditionally, feature representations have been engineered manually for most machine learning applications. Subject domain experts based on years of experience in handling and understanding the certain type of data (e.g. image, audio, speech etc.) would laboriously and intuitively engineer the features for their application. Methods to automatically learn good feature representations were always sought after, however there did not exist any universally useful method until the last decade. Since 2006, Autoencoders (AEs) and their variants have become very popular as methods to learn useful feature representations from data. This paper uses AEs within the FRC framework to learn good feature representations from data, and achieve better classification performance. A brief introduction to AE, stacked AE, stacked sparse AE, and SDSAE is made available in Appendix A.

\subsection{Naive Stacked Sparse Autoencoder based FRC}  
A naive Stacked Sparse autoencoder (SSAE) based fuzzy rule reduction method was recently suggested in our paper \cite{sevakula_sae1}. 
Network weights of SSAE were learned in unsupervised greedy layerwise fashion, and the trained SSAE was used to obtain a feature representation of fewer features. Fuzzy classification/regression models were then trained on the data with reduced features. This process is graphically depicted in Fig.\ref{sae_sample}. It is expected that the constraints applied in AE, removes redundancies in the new feature  representations. When the network is made to learn with multiple such constraints, 
the network is expected to learn the essential statistical regularities from the input. Experimentation with the new feature representation in \cite{sevakula_sae1} did give positive findings: 1) WM model \cite{modelling_wm} for regression gave similar performance, and 2) Ishibuchi et al's FRC model \cite{frc_ishibuchi2005} gave improved classification performance with fewer fuzzy rules. Overall, the naive SSAE based method \cite{sevakula_sae1} presented itself as a good starting point in this direction.

\subsection{Using Preprocessing techniques}
The approach discussed in previous subsection \cite{sevakula_sae1} has many limitations. The observed limitations are summarized below.

\begin{enumerate}
	\item Illegal gradient descent - All input features given to the SSAEs must preferably lie in the same numerical range. When the difference in input features' numerical ranges is extremely large, SSAE may even fail in the learning process. This occurs when the average activation of some hidden neurons become zero; in such a condition if the value of the input to a neuron also becomes zero, then gradient of the parameters $\mathbf{W}$ and $\mathbf{b}$ may turn into 0/0 form, also known as Not a Number (NaN) form. NaN form would naturally tamper the optimization algorithm's ability to update the parameters; thus resulting in illegal gradient descent. 
	
	\item Long time to learn - AEs may take long time to learn if the numerical ranges of input features are significantly different from each other \cite{fast_dnn}. Such phenomena are common to gradient descent based optimization techniques. 
	
	\item An AE cannot learn from nominal features, unless the input data is preprocessed or the back-propagation algorithm for learning weights is suitably modified \cite{categoricalLimitation}. 
\end{enumerate} 

The above limitations encouraged us to work on applying preprocessing methods to all input features. Some of these preprocessing methods were presented in our earlier paper \cite{sevakula_sae2}. The preprocessing functions were suggested to be similar to convex normal MFs. Having such preprocessing functions shall serve two purposes: 1) it would ensure similar numerical range for all features, and 2) it would be seen in later sections that in the presence of expert knowledge, the MFs defined in expert rules shall form the preprocessing functions.

\subsection{Need for fine-tuning techniques}
The current state-of-the-art FRCs achieved better performance than the Naive SSAE based FRC \cite{sevakula_sae1}, w.r.t. both rule reduction capability and classification accuracy. This motivated us to check if the weights of the neural network model could be further optimized to achieve better performance than the state-of-the-art FRCs. 

Once the data is preprocessed, feature representations can be easily learned using AE. Both the preprocessing step and the learning of AE are unsupervised learning tasks which do not use the class label information. Reference \cite{dl_bengio2009} states that unsupervised greedy layerwise learning of AE helps in obtaining good initialization of network weights. A third step, namely the fine-tuning step would be a supervised learning task which will fine-tune the weights such that the NN provides better support for classification. Another understanding for the same process is that, the unsupervised learning of AE helps in obtaining an initial set of high level features. However, these features are not generated/learned to give good classification performance. Hence, features are fine-tuned in a manner that will facilitate good classification performance. 

The contributions of this paper begins from hereon. The primary contributions of this paper are : 1) it provides an AE based FRC framework which incorporates pre-processing, unsupervised learning of AE, finetuning of network weights, and a facility to embed expert knowledge (described in the next two subsections), and 2) it introduces novel methods to fine-tune the network weights (described in Section III).


\begin{figure}
	\centering
	\includegraphics[scale = 0.57]{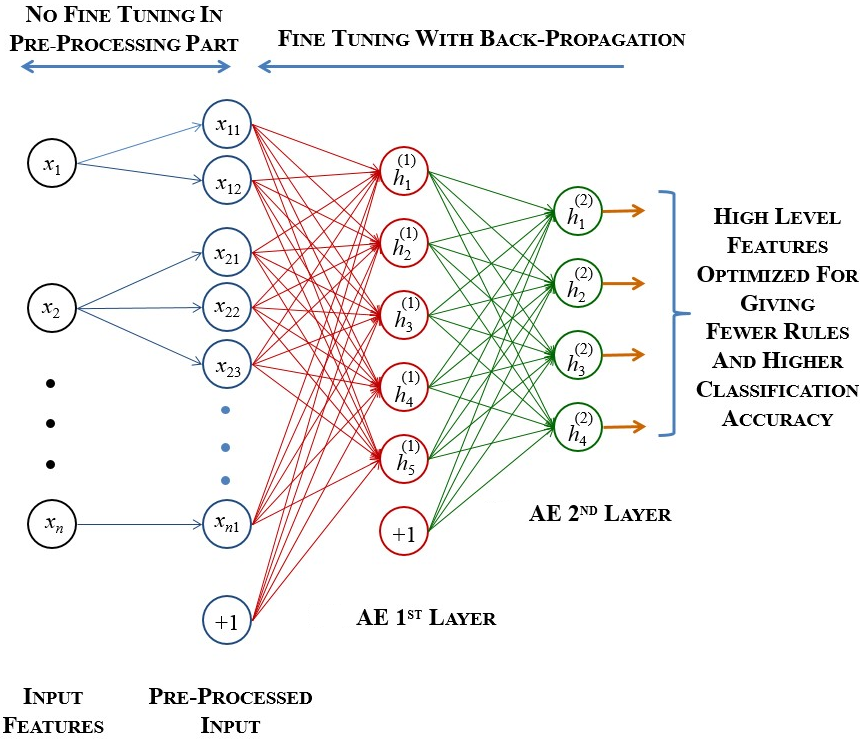} 
	\vspace{-0.8em}
	\caption{Framework for input data preprocessing, feature generation with AE, and fine tuning of network}
	\label{complete_framework}
	\vspace{-1.3em}
\end{figure} 


\subsection{Proposed AE based FRC framework}
The proposed AE based FRC framework as illustrated in Fig. \ref{complete_framework}, has 4 steps: 1) data preprocessing, 2) unsupervised learning of AE, 3) fine-tuning of network weights, and 4) training of FRC. The first step is the data-preprocessing step, where input is pre-processed and normalized with help of functions which are similar to convex membership functions. As discussed in \cite{sevakula_sae2}, by devising appropriate preprocessing functions, the number of features in preprocessed data can be more than that of input data. Additionally, since each feature may be preprocessed with a different function, the number of features after preprocessing could be different, e.g. in Fig. \ref{complete_framework}, input features $x_1$, $x_2$, $x_n$, after preprocessing give two, three and one features respectively.

The second step of the proposed framework is to pre-train the neural network. This is done through the unsupervised greedy layerwise learning of AE with the preprocessed data. As can be observed in the figure, the feature representation obtained by data at the last hidden layer of the neural network is to be used for training a FRC. Since the first two steps are unsupervised, a third step of supervised fine-tuning of the network weights is performed. Fine-tuning of weights is performed to bring out an improved feature representation which reduces the intra-class separation and increases the inter-class separation in training data samples. The fine-tuning operation typically involves a cost function which accounts the class label information of data; for our case, these cost functions would be designed to enhance classification performance of FRC. The network weights are tuned to obtain locally optimal values with the cost function. For the fourth step, training data is forward passed through the fine-tuned neural network until the last hidden layer. On performing the forward-pass, the training data would have a new feature representation at the last hidden layer. This training data with the new feature representation, along with their respective class labels is then used to train Ishibuchi et al.'s FRC based on \cite{frc_ishibuchi2005}; details regarding the FRC's \cite{frc_ishibuchi2005} training and the test procedure can be found in Appendix B. 
The preprocessing functions, the fine-tuned neural network, and the trained FRC are then stored for future use in classifying test samples.

\subsection{Embedding Expert Knowledge into the FRC}

An integral aspect of any FRC is its ability to incorporate expert knowledge into the classifier system. This subsection introduces a mechanism where, in addition to knowledge being learned from training data, existing expert knowledge would also be acknowledged, and embedded into the FRC. 
Expert knowledge is typically defined in the form of already existing rules. 
For future reference, the MFs defined in the expert rules shall be referred as expert MFs or $\text{MF}_E$ notationally.

In the absence of expert knowledge, preprocessing function parameters could be learned using various methods \cite{learning_MF,modelling_wm,sevakula_sae2}. However in the presence of expert knowledge, the expert MFs themselves are to be used as the preprocessing functions. In the equation below, input training data $\mathbf{X}$ gets preprocessed to $\mathbf{X_P}$ with membership/preprocessing function $\mathbf{PPROC}$, based on the expert MF parameters mentioned in $\mathbf{MF_E}$. 
\begin{align}
	[\mathbf{X_P}] = \text{PPROC}(\mathbf{X}, \mathbf{MF_E})  
\end{align}
        
After the training data is preprocessed, additional training samples named expert samples are introduced and made part of $\mathbf{X_P}$. The objective here is that adding expert samples would help incorporate the expert knowledge by means of having an impact on the neural network learning. 
There shall be a one-to-one correspondence between expert rules and expert samples, i.e. each rule shall generate only a single expert sample. Let $j^{th}$ feature of $\mathbf{X}$ have $K_j$ expert MFs. In such a case, $j^{th}$ feature after preprocessing with $K_j$ expert MFs would have $K_j$ output preprocessed features. Therefore, $\mathbf{X_P}$ shall have $n_p = \sum_jK_j$ features. In the antecedent part, an expert rule calls a single MF from each feature. Thus, expert samples are defined as a sparse binary vector, where features corresponding to the MFs called by the rule(s) are assigned a value of 1, and the remaining features are assigned a value of 0. Let $\mathbf{rule_g}$ refer to the $g^{th}$ rule of a total of G expert rules, then the matrix containing expert samples $\mathbf{X_E}$ is made in the following manner.                           
\begin{align}
	& \text{for}~g = 1:G  \notag\\
	& ~~~~ count = 0  \notag\\
	& ~~~~ \text{initialize vector}~ \mathbf{v} ~\text{of size}~ n_p ~\text{with all zeros}  \notag\\
	& ~~~~ \text{for}~j = 1:n_p  \notag\\
	& ~~~~ ~~~~ ind = count + l ~,~ \mathbf{rule_g}~ \text{calls} ~ l^{th} ~ \text{MF}~\text{in}~j^{th} ~\text{feature}  \notag\\	
	& ~~~~ ~~~~ v_{ind} = 1 	\notag\\
	& ~~~~ ~~~~ count = count + K_j  	\notag\\
	& ~~~~ \text{end}   \notag\\ 
	& ~~~~ \mathbf{X_E} \leftarrow \tau_g \times \{\mathbf{v}\} \notag\\ 
	& \text{end}
\end{align}

\begin{figure}
	\centering
	\includegraphics[scale = 0.28]{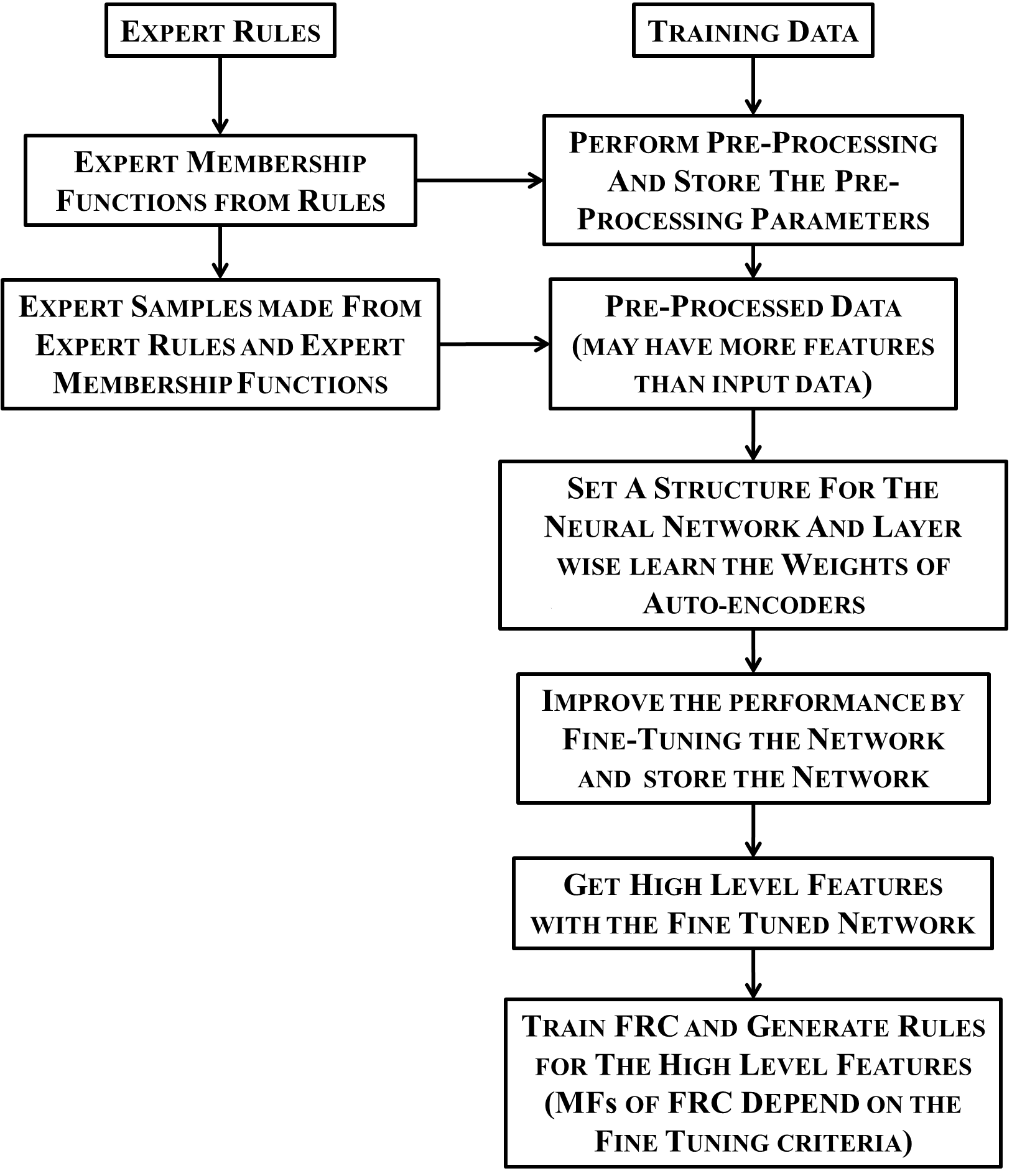}	
	\vspace{-0.5em}
	\caption{Proposed FRC framework with expert knowledge}
	\label{complete_framework_2}
	\vspace{-0.3em}
\end{figure}

When the expert sample $\mathbf{v}$ is generated for $\mathbf{rule_g}$, it is added $\tau_g$ times (upsampled $\tau_g$ times) to $\mathbf{X_E}$. Here, $\tau_g$ indicates the integer confidence/importance that the user gives to $g^{th}$ rule, which results in a total of $\sum_{g=1}^{G}\tau_g$ expert samples in $\mathbf{X_E}$. Once all $\mathbf{X_E}$ are generated, they are appended to $\mathbf{X_P}$. 
\begin{align}
\mathbf{X_P} \leftarrow  [\mathbf{X_P}, \mathbf{X_E}]
\end{align}     
The remaining steps, namely the learning of initial network weights with AEs, fine-tuning of network weights and rule generation, are the same as in the absence of expert samples. The entire process flow is summarized in Fig. \ref{complete_framework_2}. Experiments testing the effectiveness of the above approach can be found in Section IV-B where the proposed FRC is made to train on only 30\% of Iris data with and without additionally incorporating the expert rules mentioned in \cite{neuro_fugenesys}. 

Another approach for incorporating expert knowledge in the framework would be to bring the expert fuzzy rules and the fuzzy rules learned from data on a common platform, and then take a union of both types of fuzzy rules (as is done in most FRC literature). The possibility of this approach in the current framework is however left as a topic for future research.

\section{Fine Tuning of weights for FRC Modeling}

This section introduces novel fine-tuning strategies which are designed to enhance the classification performance of the proposed FRC. The proposed framework uses the FRC in \cite{frc_ishibuchi2005} as the last step to learn and classify data with the new feature representation. Reviewing the FRC in \cite{frc_ishibuchi2005} allows us to understand what is needed for FRC to give good classification performance and/or to generate minimum possible rules. Details of the algorithm for training and testing the FRC in \cite{frc_ishibuchi2005}, has been made available in Appendix B.

A pictorial presentation of the FRC on a 2D toy data is shown in Fig. 4. For simplicity, let us introduce a term called "fuzzy box" in the fuzzy rule space. The fuzzy box for a rule can be defined as the space/extent in the antecedent rule space, where the rule's influence will be most prominent as compared to all the other rules. For reference, the fuzzy boxes in Fig. 4 are shaped evenly and rectangularly.

\begin{figure}
	\centering
	\includegraphics[scale=0.4]{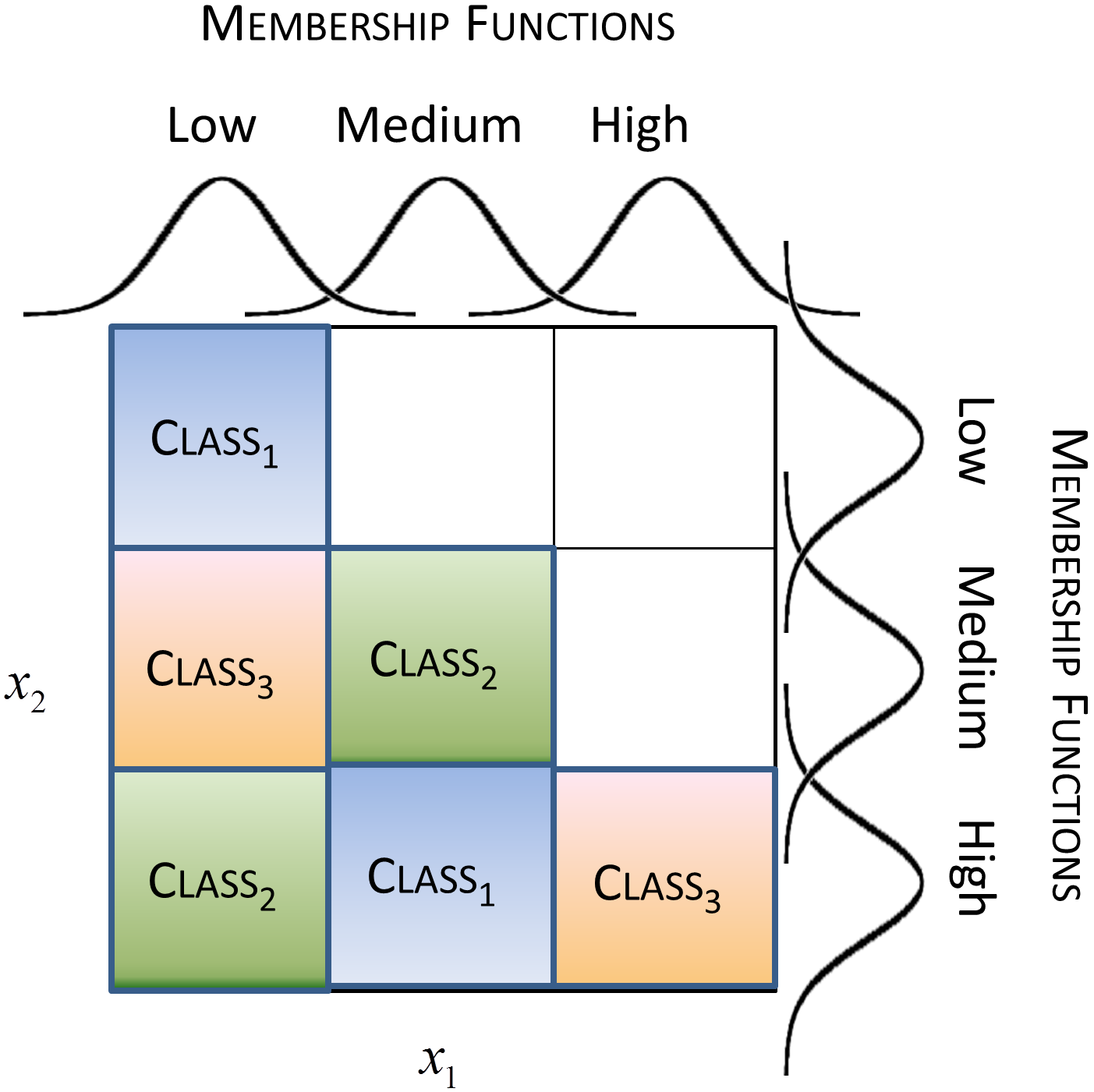}
	\label{frc_sample_fig}
	\vspace{-0.5em}
	\caption{Learning rules for FRC}
	\vspace{-3ex}
\end{figure}

The following can be inferred for improving its overall efficiency:
\begin{itemize}
	\item With $P$ unique class labels in training data, the minimum number of rules needed to model the classification problem would be $P$ rules, where each class is modeled with a single rule. For such a situation to occur, one class must be represented by only a single fuzzy box in the antecedent rule space.
	
	\item As explained in the context of bias-variance tradeoff \cite{vapnik}, good generalization is statistically achieved when the classifier achieves good classification performance on training data (high bias), with a simple model (low variance) that generally has a few model parameters. For FRC, this translates to having: 1) small training errors, 2) a small number of fuzzy rules, and 3) a small number of model parameters to model the problem. 
\end{itemize} 
Based on the above discussions, four fine-tuning procedures have been proposed here. 

\subsection{Fine Tuning I (FT-I)}
The first fine-tuning strategy aims at segregating the samples from the same class to a single fuzzy box. This is done by updating $\mathbf{W}$ and $\mathbf{b}$ in a manner such that feature values of all samples of a class converge towards their common median. By specifying the common median as a desired output value at the last hidden layer, $\mathbf{W}$ and $\mathbf{b}$ can be easily updated with the back-propagation algorithm to achieve the required objective of samples' feature values converging towards the class median. Instead of using the median, the mean of the samples could also be used as an option for fine-tuning. The median has an advantage that it is less affected by noise. The mean value is however more suitable while incorporating expert knowledge. This is because the mean value can better account for the variations in $\tau_g$ than the median. 

Ideally, only one fuzzy box per class label is expected from this strategy; hence it may be used preferably for rule reduction. A representative rule scheme expected from this strategy is shown in Fig. 5. In the figure, $\mathbf{x}^H$ refers to the sample data vector that is made available at the last hidden layer, $x^{H}_{j}$ is its value for $j^{th}$ feature, and $median_{ij}$ refers to the median (convergence point) for $i^{th}$ class and $j^{th}$ feature. In this strategy, each feature will have as many MFs as the number of classes. 

Regularization techniques involving partial addition of absolute values of weights to the cost function may be used to control $\mathbf{W}$ and $\mathbf{b}$ from taking extreme values; however their use in general is found to give suboptimal results with the FT-I finetuning technique. A primary requirement for the FT-I technique to perform well, is that feature values must appropriately converge towards their respective class' median. Regularization techniques on the other hand induce early stopping of convergence. 
Such early stopping leads to reduced interclass separation, and can increase mis-classifications in samples whose feature values are situated far from the median of their respective class. Therefore the use of regularization methods are not encouraged for FT-I.

\begin{figure}
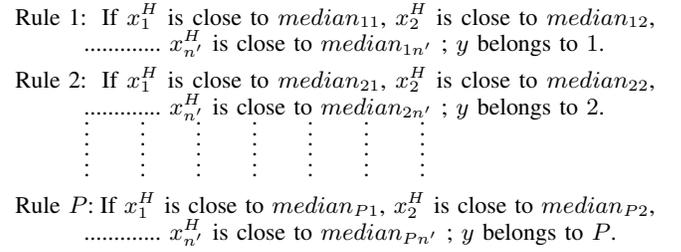

	\noindent\rule{8.8cm}{0.4pt}
    \small
	\begin{description}
		\item [Rule 1: ] $~$ If $x^H_{1}$ is close to $median_{11}$, $x^H_{2}$ is close to $median_{12}$, ............. $x^H_{n'}$ is close to $median_{1n'}$ ; $y$ belongs to $1$.\\ [-1.5ex]
		
		\item [Rule 2: ] $~$ If $x^H_{1}$ is close to $median_{21}$, $x^H_{2}$ is close to $median_{22}$, ............. $x^H_{n'}$ is close to $median_{2n'}$ ; $y$ belongs to $2$.\\ [-1.5ex]
		.$~~~~~~$.$~~~~~~$.$~~~~~~$.$~~~~~~$.$~~~~~~$.$~~~~~~$.\\ [-1.5ex]
		.$~~~~~~$.$~~~~~~$.$~~~~~~$.$~~~~~~$.$~~~~~~$.$~~~~~~$.\\ [-1.5ex]
		.$~~~~~~$.$~~~~~~$.$~~~~~~$.$~~~~~~$.$~~~~~~$.$~~~~~~$.\\ [-1.5ex]
		.$~~~~~~$.$~~~~~~$.$~~~~~~$.$~~~~~~$.$~~~~~~$.$~~~~~~$.\\ [-1.5ex]
		.$~~~~~~$.$~~~~~~$.$~~~~~~$.$~~~~~~$.$~~~~~~$.$~~~~~~$.\\ [-1.5ex]
		.$~~~~~~$.$~~~~~~$.$~~~~~~$.$~~~~~~$.$~~~~~~$.$~~~~~~$.\\ [-1.5ex]
		\item [Rule $P$: ] $~$ If $x^H_{1}$ is close to $median_{P1}$, $x^H_{2}$ is close to $median_{P2}$, ............. $x^H_{n'}$ is close to $median_{Pn'}$ ; $y$ belongs to $P$.
	\end{description}	
	\vspace{-1.5ex}
	\noindent\rule{8.8cm}{0.4pt} 
	\vspace{-3ex}
	\label{rule_scheme_ft1}
	\caption{Desired rule generation scheme for Fine Tuning 1}
	\vspace{-1em}
\end{figure}

\subsection{Training FRC}
For training an FRC \cite{frc_ishibuchi2005} over training data at the last hidden layer, it is essential to decide how the MFs would be defined for each feature. Once the MFs are defined, the FRC is expected to learn the MF parameters and rules from data. 

For our experiments with fine-tuning techniques FT-I, FT-II, FT-III and FT-IV, MFs are defined in the following manner. For each feature, as many MFs are defined as the number of classes. All MFs are convex Gaussian MFs unless specified otherwise. Accordingly, for each feature, data pertaining to the individual class labels are grouped together, and the respective MF parameters (mean and standard deviation for gaussian MFs) are learned from the grouped data. After the MF parameters are learned, the FRC mentioned in \cite{frc_ishibuchi2005} is used.

\subsection{Fine Tuning II (FT-II)}
FT-II uses Covariance Matrix Adaptation Evolution Strategy (CMA-ES) \cite{cmaes_original,cmaes_implementation} for fine-tuning the network weights and biases. CMA-ES is a popular evolution strategy which is stochastic and derivative free, and is useful for non-linear, non-convex  optimization.
The network weights and biases found after performing the unsupervised greedy layerwise learning in AE, are used as the initial values for the CMA-ES optimization. We then define the fitness function of CMA-ES as a function which optimizes the classification performance of the FRC that is built on the feature space at the last hidden layer.
When CMA-ES proceeds to optimize the weights and biases, with each new iteration, the 
feature space at the last hidden layer shall also change. Accordingly, for each iteration, the data is brought onto the new feature space, and the FRC as described in Section III-B, is trained upon the data. The fitness function to be minimized while converging upon good network weights and biases is given below. 
\begin{align}
& fitness = - T_{Acc} + \frac{G_d}{m} - \frac{P_{consequent}}{P}
\end{align}
The fitness function is based upon: 1) training accuracy of the FRC, $T_{Acc}$, 2) ratio of the number of fuzzy rules learned by the FRC $G_d$, to the total number of training samples $m$, and 3) ratio of the number of unique class labels present in the rules' consequent part $P_{consequent}$ to the total number of unique class labels present in training data $P$. The fitness function aims to have high training accuracy, have few generated rules by the FRC, and ensure that the generated rules have covered all $P$ class labels. In the case of imbalanced datasets, an objective function that focuses on high training accuracy and few fuzzy rules, can allow the FRC to miss out rules for the class labels that have few corresponding training samples. To avoid the situation where not a single rule is generated for few class labels (as consequent part), the third term of the fitness function penalizes the complete absence of class label(s) in fuzzy rules. 

For huge networks, the number of network weights and biases to be optimized can be very large. Furthermore, in each generation of CMA-ES, a single FRC is to be trained. On these accounts, the FT-II approach though effective, is computationally very expensive.

\subsection{Fine Tuning III (FT-III)}
As we have already explained, FT-I strategy tunes the network weights and biases in a manner where feature values at the last hidden layer converge towards their common median of each class. Such convergence is found to provide good results in many situations. However, in few cases the class medians could be very close to each other, and may not provide enough separability across classes in the new feature space. FT-III uses CMA-ES to identify the optimal points for convergence.
Taking the class medians as the initial convergence points, CMA-ES iteratively updates the convergence points based on the same fitness function (4) as used in FT-II.

For each update, i.e. each generation of the convergence points by CMA-ES, the networks weights and biases are reset to the values obtained by unsupervised greedy layerwise learning of AE. The weights and biases are then updated through the back-propagation algorithm such that feature values at the last hidden layer converge towards the updated convergence points. Finally, FRC is trained on the data, and the value of the fitness function is computed. 

Each feature in the new feature space has $P$ convergence points. The total number of the convergence points would equal $n\times P$. Typically, this number would be much smaller than the number of network weights and biases. Therefore, the number of generations and updates required by CMA-ES is expected to be much fewer. In spite of the fact that each update requires FRC training and network weights updation, FT-III is computationally less expensive than FT-II, and provides comparably good classification performance.

\subsection{Fine Tuning IV (FT-IV)}
FT-IV strategy also aims at coming up with the optimal convergence points. The difference however is that FT-IV does not use CMA-ES for optimizing the convergence points, instead it defines a differentiable objective/cost function to obtain the convergence points. The cost function to be optimized is shown below.
\begin{align}
\argminA_\mathbf{C} &~ \frac{1}{2}\sum_{j=1}^{P} \sum_{i, y_i = j}^{m} \frac{\left(\mathbf{x}_i^{H} - \mathbf{C}_j\right)^{2}}{m_j} \notag \\
& - ~ \frac{\beta}{2(P-1)}\sum_{j=1}^{P} \sum_{l,l\neq j}^{P} \left(\mathbf{C}_j - \mathbf{C}_l \right)^2 + \frac{\zeta}{2P}\sum_{j=1}^{P}\mathbf{C}_j^2
\end{align}

In the cost function above, $\mathbf{x}_i^H$ refers to the feature vector of $i^{th}$ sample as obtained at the last hidden layer, $\mathbf{C}_j$ refers to the vector containing the convergence points of all features at the last hidden layer for $j^{th}$ class, and $m_j$ refers to the number of samples belonging to $j^{th}$ class label. The cost function aims at: 1) obtaining a small intra-class distance between the samples of each class and their convergence point, with the first term, and 2) obtaining high separation between inter-class convergence points, with the second term. It may be noted that both terms are normalized by dividing them, with either the number of samples or the number of classes, across which the summation takes place. The variable $\beta$ helps in deciding the tradeoff as to which term is more significant while deciding the convergence points. To avoid the case where the convergence points approach to extreme values during the optimization, a third term $\frac{\zeta}{P}\sum_{j=1}^{P}\mathbf{C}_j^2$ is included as a regularization term.

Since the cost function is differentiable, gradient descent based optimization techniques could be used to arrive at the locally optimal convergence points. On these accounts, FT-IV is much quicker than FT-III in deciding the optimal convergence points. To use standard optimization packages, equations for computing the gradients are generally required. The equation for computing gradients w.r.t. the convergence points of each class is shown below. 
\begin{align} 
\nabla \mathbf{C}_j = \sum_{\forall i, y_i = j}^{m_j} \frac{\left(\mathbf{C}_j - \mathbf{x}_i^{H}\right)}{m_j} & - \sum_{\forall l, l \neq j}^{P} \frac{\beta \left(\mathbf{C}_j - \mathbf{C}_l \right)}{P-1} + \zeta\sum_{j=1}^{P} \frac{\mathbf{C}_j}{P}
\end{align} 

It may be noted that $\beta$ should not have a large value. This is because in the cost function if $\frac{\beta}{2(P-1)}\sum_{j=1}^{P} \sum_{l,l\neq j}^{P} \left(\mathbf{C}_j - \mathbf{C}_l \right)^2$ is larger than the other two terms, then the convergence points would be optimized in a manner where the cost function is $-\infty$.


\section{Experimentation and Results}

Experiments have been carried over 5 real world datasets from UCI repository \cite{uci} as detailed in Table \ref{datasets}. The main motivations for choosing these five datasets are : 1) easy comparison with earlier FRC literature, and 2) the computationally expensive fine-tuning procedure of FT-II. Considering that many FRC studies have evaluated their performance over 10-fold cross-validation, this paper also does evaluation on 10-fold cross-validation. 
Test performance in terms of the classification accuracy and the number of generated rules for FRC, were evaluated for all 10 pairs of training and test data, then averaged and noted down. The cross validation sets used are made available online at ``http://iitk.ac.in/idea/datasets/sdsaefrc".

As mentioned in the Introduction section, we have used the SDSAE implementation of AEs for all our experiments in this paper. A limited-memory BFGS optimization algorithm of the minFunc package \cite{minfunc_kent} in MATLAB was used for all optimization during the learning of weights and biases in SDSAE and FFN. The stopping criterion used for optimization was when either the tolerance becomes less than 1e-6 or the number of iterations reaches the upper limit of 400 epochs. For the CMA-ES based optimization in FT-II and FT-III, Hansen et al.'s CMA-ES \cite{cmaes_implementation} was used. 
For FT-IV, the value of $\zeta$ was kept at 0.05, and $\beta$ was the largest value among \{0.1,0.2,...,1.0\} such that none of the resultant convergence points had an absolute value larger than 10.


\begin{table}[t]
	\centering
	\caption{Details of Experimented Datasets}
	\begin{tabular}{c|c|c|c} \\

		\hline \hline
		Datasets & Attributes & Samples & Classes \\ 
		\hline
		Iris & 4 & 150 & 3  \\
		Wine & 13 & 178 & 3  \\
		Breast Cancer & 9 & 683 & 2  \\
		Sonar & 60 & 208 & 2 \\
		Pima & 8 & 768 & 2  \\
		\hline \hline
	\end{tabular}
	\label{datasets}
	\vspace{-0.5em}
\end{table}


\begin{figure}
	\centering
	\includegraphics[scale=0.3]{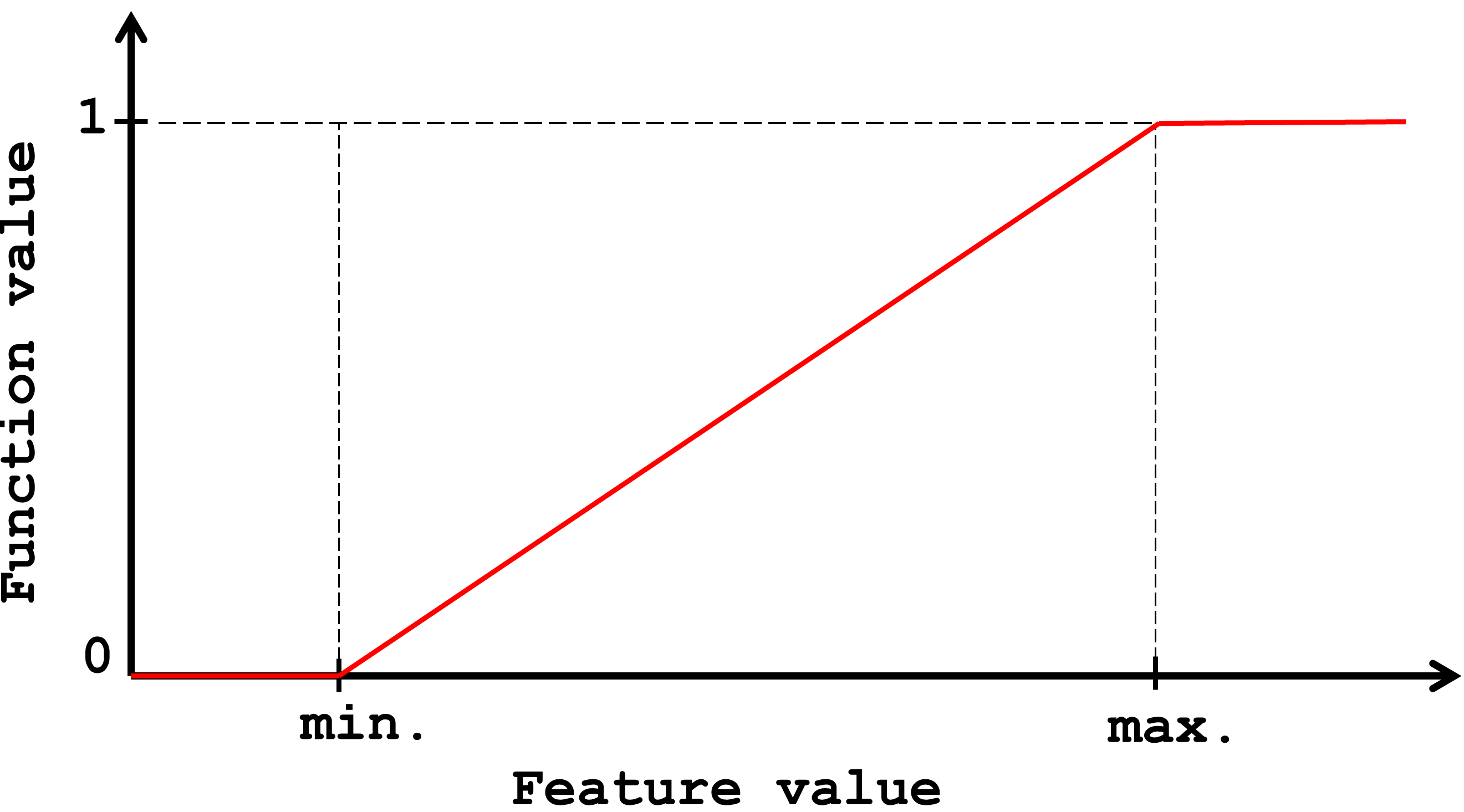}
	\label{zeroone_mf}
	\caption{Preprocessing function used for all experiments}
	\vspace{-3ex}
\end{figure}


\begin{table*}
	\centering
	\caption[caption]{Detailed Results with Proposed Methods (in \%)}
    \vspace{-2em}
	\begin{tabular}{*{1}{>{\centering\arraybackslash\rule{0pt}{0.50em}}m{1.5cm}} | c |
			*{1}{>{\centering\arraybackslash\rule{0pt}{0.50em}}m{1.7cm}} 
			*{1}{>{\centering\arraybackslash\rule{0pt}{0.50em}}m{1.7cm}} 
			*{1}{>{\centering\arraybackslash\rule{0pt}{0.50em}}m{1.7cm}}
            *{1}{>{\centering\arraybackslash\rule{0pt}{0.50em}}m{1.7cm}}}\\		
		\hline \hline
		
		Dataset & Param & FT-I & FT-II & FT-III & FT-IV \\ [0.5ex]

		\hline
		\multirow{3}{*}{Iris}  & Accuracy & 97.33\rpm3.22    & 98.00\rpm4.50  & 96.67\rpm5.67  & 98.00\rpm3.22  \\[0.5ex]
		
		& Rule Size , $\rho$ & 3.0, 0.3  & 5.9, 0.2  & 5.2, 0.4  & 5.1, 0.9\\[-0.3ex]
		
		& Net. Arch. &  (4,4)  &  (4,4)  & (4,4) & (4,4)   \\[1.5ex] 
		
		
		\multirow{3}{*}{Wine}  & Accuracy & 96.60\rpm4.81  &96.57\rpm6.22   &  97.22\rpm4.72  & 96.08\rpm4.59\\[0.5ex]
		
		&  Rule Size , $\rho$  & 3.2, 0.4  & 3.5, 0.7  & 9.2, 0.2  &  4.7, 0.5\\[-0.3ex]
		
		& Net. Arch. & (13,11,9)  & (11,9)  & (11,9) & (11,9) \\[1.5ex]

		
		\multirow{3}{*}{\begin{tabular}[x]{@{}c@{}}Breast\\Cancer\end{tabular}}   & Accuracy & 97.36\rpm1.5    & 97.07\rpm1.55  & 97.36\rpm2.17 & 97.80\rpm0.78  \\[0.5ex] 
		
		&  Rule Size , $\rho$ & 2, 0.2   & 2.1, 0.2  &  3.4,0.1  & 5.3, 0.9  \\[-0.3ex]
		
		&  Net. Arch.  & (9,5,3)  & (9,5,3)  & (9,5,3)  &  (9,5,3) \\ [1ex]

		
		\multirow{3}{*}{Sonar}  & Accuracy & 83.64\rpm11.11  &  74.43\rpm12.65 & 79.26\rpm5.85 & 78.81\rpm6.17\\[0.5ex]
		
		&  Rule Size , $\rho$ & 9.8, 0.7  &   45.3, 0.9  & 5, 0.1  & 2.3, 0.8\\[-0.3ex]
		
		& Net. Arch.  & (60,50,40)  & (60,30,4)  & (60,50,40)  & (60,50,40) \\[1.5ex] 
		
		
		\multirow{3}{*}{Pima}  & Accuracy & 68.23 \rpm6.57  & 76.81\rpm3.98 & 77.07\rpm4.73  &  78.38\rpm10.09\\[0.5ex]

		&  Rule Size , $\rho$ & 25.5, 0.2 & 8.1, 0.2 & 9,0.8 & 11.3, 0.8\\[-0.3ex]
		
		& Net. Arch.  & (8,4)  & (8,4)  & (8,4) & (8,4) \\[1.5ex]

		\hline \hline
	\end{tabular}
	\label{results_comp}
    \vspace{-1.5em}
\end{table*}


This paper presented four fine-tuning strategies in the FRC framework. To compare them across each other, independent experiments were performed with each of the fine-tuning strategies. For simplicity, the data in all experiments was preprocessed with a preprocessing function that is similar to the zero-one (also known as "max-min") normalization. The preprocessing function is shown in Fig. 6. In the figure, "min." and "max." values refer to the minimum and maximum values of each feature in the training data.

For the experiments, the network weights and biases were initialized with unsupervised greedy layerwise learning of SDSAE. The weights, biases were then fine-tuned with each of the proposed fine-tuning procedures. Finally the FRC mentioned in Section III.B was trained on data obtained at the last hidden layer. The results for all experiments have been tabulated in Table \ref{results_comp}. For each dataset in the table, the first row presents the average classification accuracy values (Accuracy), the second row presents the average number of rules generated by the FRC (Rule Size) and the sparsity parameter value $\rho$ that gave these optimal results. The third row mentions the network architecture (Net. Arch.) by mentioning the number of units present in the input and hidden layer(s). 
The network architecture was intuitively selected, and multiple architectures were not tested and compared. Bengio \cite{dl_bengio2012} mentions that as long as the network has sufficient flexibility to model the non-linear relationships between features, the performance of networks does not depend very strongly on network architecture. Deciding the optimal network architecture for the proposed FRC, is beyond the scope of this paper. 
In general, an appropriate specification of 
$\rho$ without prior knowledge is difficult. Moreover, as networks become deeper, the resultant features and classification performance can be significantly affected by the value chosen for $\rho$. To find a reasonable value, $\rho$ was varied across \{0.1,0.2,.....,0.9\} and the value which gave the best average classification performance was chosen. It may be noted that during the layerwise training of AEs \cite{sevakula_sae1,dl_bengio2009,dl_ng_lecture_notes}, white noise of 10dB was added to the input of AEs for incorporating the denoising constraint \cite{dl_denoising_ae}.


\begin{table}[t]
	\centering
	\caption{Comparing average test errors (in \%)}
    \vspace{-1.0em}
    \begin{tabular}{*{1}{>{\centering\arraybackslash\rule{0pt}{0.50em}}m{2.7cm}} |
			*{1}{>{\centering\arraybackslash\rule{0pt}{0.50em}}m{0.4cm}} |
			*{1}{>{\centering\arraybackslash\rule{0pt}{0.50em}}m{0.4cm}} |
			*{1}{>{\centering\arraybackslash\rule{0pt}{0.50em}}m{0.6cm}} |
            *{1}{>{\centering\arraybackslash\rule{0pt}{0.50em}}m{0.5cm}} |
            *{1}{>{\centering\arraybackslash\rule{0pt}{0.50em}}m{0.5cm}} 
            }\\	

		\hline \hline
		Reference   & Iris & Wine & Cancer & Sonar  & Pima  \\ \hline
		Quinlan \cite{comparison_quinlan1996}   & 4.80 & - & 5.26 & 25.60  & 25.40 	 \\
		Elomaa \cite{comparison_elomaa1999}     & 6.60 & 5.60 & 5.30 & 25.10 & 25.70	 \\
		Sanchez \cite{comparison_sanchez2001}   & -  & - & 4.65 & - & -   \\
		Slave \cite{comparison_slave}           & 4.28 & 6.20 & - & -  & - 	\\
        Slave\_v0 \cite{slave_overview}	        & 4.00 & 6.18 & 6.30 & 30.26 & 23.82  \\
		Ishibuchi \cite{frc_ishibuchi_comparison}   & - & 6.90 & 3.25 & 24.06 & - 	 \\
		Abonyi \cite{comparison_abonyi2003}     & 3.89 & 8.78 & 3.18 & -  & 26.95 	\\
		Guan \cite{comparison_guan2004}         & 4.40 & 8.33 & 4.70 & -  & - 	\\
		Ishibuchi \cite{frc_ishibuchi_gbml}     & 5.33  & 4.94 & 3.32 & 23.70 & 24.17  \\
		SGERD \cite{reduction_sgerd}            & 3.07 & 3.81 & 2.98 & 22.80 & 25.36  \\
		FURIA \cite{furia_c}                    & 4.67 & 2.85  & 4.69 & 20.24 & 26.17  \\
		ChiRW \cite{cordon, frc_ishibuchi2005}  & 7.34 & 6.18  & 7.92 & 39.95 & 26.93  \\
		WF \cite{wfc_nakashima}                 & 4.00 & - & 2.93 & - & 26.29  \\
		FARC-HD \cite{FARCHDC_herrera}          & 4.00 & 5.65 & - & 19.81 & 24.34  \\
        GFS-Adaboost \cite{gfs_adaboost}        & 4.00 & 11.80 & 5.28 & 53.38 & 25.01  \\
        MaxLogitBoost \cite{gfs_logitboost}     & 7.33 & 5.59 & 4.39 & 53.38 & 24.86  \\
        IVTURS \cite{ivturs}                    & 3.33 & 7.39 & 3.22 & 21.71 & 25.63  \\
        SDSAE+Softmax\cite{dl_ng_lecture_notes} & 4.00 & 2.81 & 6.90 & 16.86 & 24.52  \\  
        \textbf{Proposed}-(overall)    &  \textbf{2.00} & \textbf{2.78}  &  \textbf{2.20} &  \textbf{16.38} & \textbf{21.62}  \\  
		\textbf{Proposed}-(FT-I)    &  2.67 			& 3.4  			&  2.64 &  \textbf{16.38} &  31.77 \\
        \textbf{Proposed}-(FT-II) & \textbf{2.00} 		& 3.43 			& 2.93 			 & 25.57 	& 23.19  \\ 	
        \textbf{Proposed}-(FT-III)  & 3.33	& \textbf{2.78} & 2.64  & 20.74  			& 22.93  \\
        \textbf{Proposed}-(FT-IV)  & \textbf{2.00}  & 3.92 & \textbf{2.20} & 21.19 & \textbf{21.62}  \\
		\hline \hline
	\end{tabular}
	\label{classification_comp}
	\vspace{-1.0em}
\end{table}


Comparison of classification performance across various studies on FRC has been presented in Table \ref{classification_comp}. Results for \cite{comparison_quinlan1996,comparison_elomaa1999,comparison_sanchez2001,comparison_slave,slave_overview,frc_ishibuchi_comparison,comparison_abonyi2003,comparison_guan2004,frc_ishibuchi_gbml,reduction_sgerd}
were directly taken from the papers, as they had observed a similar experimentation procedure of measuring the performance over 10-fold cross validation. Results for \cite{furia_c, cordon, frc_ishibuchi2005, wfc_nakashima, FARCHDC_herrera, gfs_adaboost, gfs_logitboost, ivturs} were generated using KEEL software \cite{keel}. Results were also generated with unsupervised SDSAE followed by Softmax fine-tuning \cite{dl_ng_lecture_notes} ("SDSAE+Softmax"), which is a very popular deep learning classification framework. Results for this case were also measured over 10-fold cross-validation, and they were generated for sparsity parameter values ranging from \{0.1, 0.2, ..., 0.9\}. The best test-errors have been noted down in the table. 
The row for "Proposed - (overall)" presents the overall best classification performance achieved by the four proposed methods. The last 4 rows of the table show the best test error rates with each of the fine-tuning strategies. On comparing the results, 
it is clear that the proposed FRC provides the best average test error for all the five datasets.

FT-I is the procedure of fine-tuning the weights and biases in a manner where feature values at the last hidden layer converge to values (class median for FT-I) that are common to all samples of each class. FT-III and FT-IV have the same theme, and optimize the convergence points by means of a proposed fitness and cost function respectively. We can observe from Table III, that such a theme is very effective in obtaining good classification results. FT-I, FT-III, and FT-IV together obtained the best classification results across all the compared FRC methods, and these results were even better than a popular deep learning classification framework \cite{dl_ng_lecture_notes}.

While FT-II and FT-III obtained reasonably good classification performance, they were found to be computationally expensive. The heavy computation is expected because: 1) CMA-ES is iteratively invoked to optimize parameters at a time, 2) In FT-II, there are too many parameters to optimize and this number can become really huge with larger networks, and 3) computing the fitness function in each iteration of CMA-ES, involves multiple steps of computations. These multiple steps include a forward pass of data till the last hidden layer, training of a FRC with data obtained at the last hidden layer, and lastly in the case of FT-III, weights and bias are updated through the back-propagation algorithm. Due to the high computational expense, it was difficult to test and generate results for larger datasets with FT-II and FT-III, especially because they needed larger networks.

To perform statistical comparisons with existing FRCs, and to establish the effectiveness of our proposed approach, we performed experiments with 10 additional datasets. Keeping the computational requirements in view, these additional experiments were performed with only FT-I and FT-IV strategies, and they were compared with four popular FRCs, namely Slave\_v0\cite{slave_overview}, SGERD\cite{reduction_sgerd}, ChiRW\cite{cordon, frc_ishibuchi2005}, and IVTURS\cite{ivturs}. The additional results, and the statistical tests made across the 15 datasets (5 + 10 additional), have been detailed in Appendix D. Our findings from the statistical tests are as follows: 1) Using the Friedman test \cite{demsar}, the Null hypothesis is rejected at $\alpha=0.05$ (confidence level of 95\%), 2) Using the Bonferroni-Dunn posthoc test \cite{demsar}, it is ascertained with $\alpha=0.05$, that proposed methods provide statistically better results than IVTURS, Slave\_v0, SGERD, and ChiRW, and 3) Using the Wilcoxon Signed-Ranks test \cite{demsar} for pairwise comparison, it is ascertained with $\alpha=0.05$, that proposed method provide statistically better results than IVTURS, Slave\_v0, SGERD, and ChiRW. We finally wish to point out that, the proposed FRC with FT-I/FT-IV strategy is empirically much quicker in execution than the Slave\_v0 and the IVTURS, and is comparable to the SGERD in execution time. We also wish to acknowledge that, the proposed approach with FT-I and FT-IV strategy, works especially well on datasets having few classes (two or three classes preferably).

Typically, DNNs are used for training on large data. Another important point for discussion is the feasibility of using deep networks on small datasets here, for achieving good FRC performance. It can be observed from Table \ref{results_comp}, that the network architecture used on all these datasets offer sufficient constraints. The constraints are: 1) the original input space is made to be represented by fewer features, 2) there is a sparsity constraint in hidden nodes, and 3) the denoising criterion is included while training each autoencoder. The quality, with which network weights are learned and tuned, is dependent on two factors, namely the size of data and the constraints with which they are to be learned. In the experiments, we tried to handle the limitation of small data size by using increased constraints on network architecture, sparsity in hidden layers and denoising ability of AEs. 

\section{Extensions}
The proposed FRC, like other FRCs allows the incorporation of expert knowledge. This section discusses the effectiveness of incorporating expert knowledge into the proposed framework, and also discusses the limitations and possible future efforts that could be made to further improve the proposed FRC. 

\subsection{Incorporating expert knowledge}
The Iris dataset is a popular dataset that many FRCs have experimented with, and unlike many datasets, it was easy to obtain its expert rules. Five expert rules as defined in \cite{neuro_fugenesys} were used for our experiments; details regarding the rules, the MFs and expert samples retrieved, are shown in Appendix C. 

For experimentation, 30\% of data i.e. first 15 samples from each of the three classes i.e. 45 samples were selected for training the classifiers, and their classification performance was measured across all 150 samples. Three FRCs were trained for comparison: 1) Proposed FRC with FT-I fine-tuning but without expert knowledge, 2) Training a neural network with SDSAE followed by Softmax fine-tuning \cite{dl_ng_lecture_notes}, 3) Proposed FRC that uses the expert MFs for preprocessing, FT-I for fine-tuning but does not include the expert samples during training, and, 4) Proposed FRC that uses the expert MFs for preprocessing, FT-I for fine-tuning and uses the expert samples during training with $\tau_g$ of 9 for each expert sample. The neural network architecture used for all four cases was the same as in Table II. Results were found with sparsity parameter values ranging across \{0.1,0.2,...,0.9\}, and the best classification performances is shown in Table \ref{expert_performance}. The results indicate a progressive increase in performance with 1) inclusion of the expert MFs in preprocessing and 2) inclusion of the expert samples in training. In other words, the method for incorporating expert knowledge seems to work as expected.


\begin{table}[t]  
	\centering
	\caption{Performance of proposed FRC on Iris data with and without expert knowledge}
	\vspace{-0.5em}
	\begin{tabular}{
			*{1}{>{\centering\arraybackslash\rule{0pt}{0.50em}}m{1.5cm}}|*{1}{>{\centering\arraybackslash\rule{0pt}{0.50em}}m{1.5cm}} |*{1}{>{\centering\arraybackslash\rule{0pt}{0.50em}}m{1.5cm}} |*{1}{>{\centering\arraybackslash\rule{0pt}{0.50em}}m{2cm}} } 
		\hline \hline
		FRC without Expert Knowledge & SDSAE + Softmax & Expert MFs + FRC & Expert MFs + Expert Samples + FRC\\
		\hline
		
		85.33 \% & 95.33 \% & 96.00 \% & \textbf{96.92} \% \\
		
		\hline \hline
	\end{tabular}
	\label{expert_performance}
	\vspace{-1.5em}
\end{table}


\subsection{Possible Future work}
It may be noted that the proposed framework does not use any form of feature selection or rule selection (as they are beyond the scope of this paper) which most FRC frameworks use for obtaining good results. Additionally, a very basic preprocessing function was used here for all experiments. For future work, feature selection, rule selection, using better preprocessing MFs, and MF tuning \cite{fuzzyhmm} strategies could be included for improving this framework. Designing fine-tuning strategies which can obtain even better convergence points could be another future work. The paper also does not address the concern as to how one may interpret the rules generated by the network, and this is an important future research topic.

\section{Conclusions} 

A framework that brings the advantages of both AEs and FRC can be a great asset for classification systems. This paper has successfully managed to introduce a well performing, introductory framework for AE based FRCs. The framework not only managed to learn the complex non-linear relationship amongst features, but also provided flexibilty to incorporate existing expert knowledge into the system. To improve the classification performance, the paper introduced novel fine-tuning techniques. The introduced methods were experimented on five benchmark datasets, and the methods were found to give best results among the examined FRC methods. Additionally, the theme of designing fine-tuning strategies in a way where feature values at the last hidden layer converge to common values for each class, appears to be very effective. 
Future work could include the use of feature selection, rule selection, MF tuning strategies to make this framework efficient for larger datasets. Another important future work could be on how users can interpret the rules generated by the FRC framework.

\bibliographystyle{IEEEbib}
\bibliography{strings,refs}

\begin{thebibliography}{10}

\bibitem{smc_howgoodarefrc}
L.~I. Kuncheva,
\newblock ``How good are fuzzy if-then classifiers?,''
\newblock {\em IEEE Trans. Syst., Man, Cybern., Part B}, vol. 30, no. 4, pp.
  501--509, 2000.

\bibitem{cordon}
O.~Cord{\'o}n, M.~J. del Jesus, and F.~Herrera,
\newblock ``A proposal on reasoning methods in fuzzy rule-based classification
  systems,''
\newblock {\em International Journal of Approximate Reasoning}, vol. 20, no. 1,
  pp. 21 -- 45, 1999.

\bibitem{frc_ishibuchi1992}
H.~Ishibuchi, K.~Nozaki, and H.~Tanaka,
\newblock ``Distributed representation of fuzzy rules and its application to
  pattern classification,''
\newblock {\em Fuzzy sets and systems}, vol. 52, no. 1, pp. 21--32, 1992.

\bibitem{frc_ishibuchi1995}
H.~Ishibuchi, K.~Nozaki, N.~Yamamoto, and H.~Tanaka,
\newblock ``Selecting fuzzy if-then rules for classification problems using
  genetic algorithms,''
\newblock {\em IEEE Trans. Fuzzy Syst.}, vol. 3, no. 3, pp. 260--270, 1995.

\bibitem{modelling_wm}
L.~X. Wang and J.~M. Mendel,
\newblock ``Generating fuzzy rules by learning from examples,''
\newblock {\em IEEE Trans. Syst., Man, Cybern.}, vol. 22, no. 6, pp.
  1414--1427, 1992.

\bibitem{tnnls_mvp}
R.~K. Sevakula and N.~K. Verma,
\newblock ``Assessing generalization ability of majority vote point
  classifiers,''
\newblock {\em IEEE Trans. Neural Netw. and Learn. Syst.}, 2016.

\bibitem{fs_chen}
Y.~C. Chen, N.~R. Pal, I.~Chung, et~al.,
\newblock ``An integrated mechanism for feature selection and fuzzy rule
  extraction for classification,''
\newblock {\em IEEE Trans. Fuzzy Syst.}, vol. 20, no. 4, pp. 683--698, 2012.

\bibitem{reduction_jin}
Y.~Jin,
\newblock ``Fuzzy modeling of high-dimensional systems: complexity reduction
  and interpretability improvement,''
\newblock {\em IEEE Trans. Fuzzy Syst.}, vol. 8, no. 2, pp. 212--221, 2000.

\bibitem{reduction_taniguchi}
T.~Taniguchi, K.~Tanaka, H.~Ohtake, and H.~O. Wang,
\newblock ``Model construction, rule reduction, and robust compensation for
  generalized form of takagi-sugeno fuzzy systems,''
\newblock {\em IEEE Trans. Fuzzy Syst.}, vol. 9, no. 4, pp. 525--538, 2001.

\bibitem{reduction_luo}
M.~Luo, F.~Sun, and H.~Liu,
\newblock ``Hierarchical structured sparse representation for t-s fuzzy systems
  identification,''
\newblock {\em IEEE Trans. Fuzzy Syst.}, vol. 21, no. 6, pp. 1032--1043, 2013.

\bibitem{reduction_sgerd}
E.~G. Mansoori, M.~J. Zolghadri, and S.~D. Katebi,
\newblock ``Sgerd: A steady-state genetic algorithm for extracting fuzzy
  classification rules from data,''
\newblock {\em IEEE Trans. Fuzzy Syst.}, vol. 16, no. 4, pp. 1061--1071, 2008.

\bibitem{frc_ishibuchi2004}
H.~Ishibuchi and T.~Yamamoto,
\newblock ``Fuzzy rule selection by multi-objective genetic local search
  algorithms and rule evaluation measures in data mining,''
\newblock {\em Fuzzy sets and systems}, vol. 141, no. 1, pp. 59--88, 2004.

\bibitem{smc_perf_evaluation}
H.~Ishibuchi, T.~Nakashima, and T.~Murata,
\newblock ``Performance evaluation of fuzzy classifier systems for
  multidimensional pattern classification problems,''
\newblock {\em IEEE Trans. Syst., Man, Cybern., Part B}, vol. 29, no. 5, pp.
  601--618, 1999.

\bibitem{ga_review}
M.~Fazzolari, R.~Alcala, Y.~Nojima, H.~Ishibuchi, and F.~Herrera,
\newblock ``A review of the application of multiobjective evolutionary fuzzy
  systems: Current status and further directions,''
\newblock {\em IEEE Trans. Fuzzy Syst.}, vol. 21, no. 1, pp. 45--65, 2013.

\bibitem{fs_muni_genetic}
D.~P. Muni, N.~R. Pal, and J.~Das,
\newblock ``Genetic programming for simultaneous feature selection and
  classifier design,''
\newblock {\em IEEE Trans. Syst., Man, Cybern. Part B}, vol. 36, no. 1, pp.
  106--117, 2006.

\bibitem{smc_featureselection}
H.~M. Lee, C.~M. Chen, J.~M. Chen, and Y.~L. Jou,
\newblock ``An efficient fuzzy classifier with feature selection based on fuzzy
  entropy,''
\newblock {\em IEEE Trans. Syst., Man, Cybern., Part B}, vol. 31, no. 3, pp.
  426--432, 2001.

\bibitem{reduction_yam}
Y.~Yam, P.~Baranyi, and C.~T. Yang,
\newblock ``Reduction of fuzzy rule base via singular value decomposition,''
\newblock {\em IEEE Trans. Fuzzy Syst.}, vol. 7, no. 2, pp. 120--132, 1999.

\bibitem{reduction_setnes}
M.~Setnes and R.~Babu{\v{s}}ka,
\newblock ``Rule base reduction: some comments on the use of orthogonal
  transforms,''
\newblock {\em IEEE Trans. Syst., Man, Cybern. Part C}, vol. 31, no. 2, pp.
  199--206, 2001.

\bibitem{reduction_tao}
C.~W. Tao,
\newblock ``A reduction approach for fuzzy rule bases of fuzzy controllers,''
\newblock {\em IEEE Trans. Syst., Man, Cybern. Part B}, vol. 32, no. 5, pp.
  668--675, 2002.

\bibitem{reduction_koczy}
L.~T. Koczy and K.~Hirota,
\newblock ``Size reduction by interpolation in fuzzy rule bases,''
\newblock {\em IEEE Trans. Syst., Man, Cybern. Part B}, vol. 27, no. 1, pp.
  14--25, 1997.

\bibitem{frc_ishibuchi2001}
H.~Ishibuchi and T.~Nakashima,
\newblock ``Effect of rule weights in fuzzy rule-based classification
  systems,''
\newblock {\em IEEE Trans. Fuzzy Syst.}, vol. 9, no. 4, pp. 506--515, 2001.

\bibitem{frc_ishibuchi2005}
H.~Ishibuchi and T.~Yamamoto,
\newblock ``Rule weight specification in fuzzy rule-based classification
  systems,''
\newblock {\em IEEE Trans. Fuzzy Syst.}, vol. 13, no. 4, pp. 428--435, 2005.

\bibitem{neuro_SuPFuNIS}
S.~Paul and S.~Kumar,
\newblock ``Subsethood-product fuzzy neural inference system (supfunis),''
\newblock {\em IEEE Trans. Neural Netw.}, vol. 13, no. 3, pp. 578--599, 2002.

\bibitem{neuro_fugenesys}
M.~Russo,
\newblock ``Fugenesys-a fuzzy genetic neural system for fuzzy modeling,''
\newblock {\em IEEE Trans. Fuzzy Syst.}, vol. 6, no. 3, pp. 373--388, 1998.

\bibitem{neuro_ctlin}
M.~F. Han, C.~T. Lin, and J.~Y. Chang,
\newblock ``A compensatory neurofuzzy system with online constructing and
  parameter learning,''
\newblock in {\em IEEE Int. Conf. Systems Man and Cybernetics (SMC'10)}, 2010,
  pp. 552--556.

\bibitem{neuro_debrup}
D.~Chakraborty and N.~R. Pal,
\newblock ``A neuro-fuzzy scheme for simultaneous feature selection and fuzzy
  rule-based classification,''
\newblock {\em IEEE Trans. Neural Netw.}, vol. 15, no. 1, pp. 110--123, 2004.

\bibitem{neuro_debrup_connectionist}
D.~Chakraborty and N.~R. Pal,
\newblock ``Selecting useful groups of features in a connectionist framework,''
\newblock {\em IEEE Trans. Neural Netw.}, vol. 19, no. 3, pp. 381--396, 2008.

\bibitem{dl_bengio2009}
Y.~Bengio,
\newblock ``Learning deep architectures for ai,''
\newblock {\em Foundations and trends{\textregistered} in Machine Learning},
  vol. 2, no. 1, pp. 1--127, 2009.

\bibitem{deepCascadeFRC}
P.~Angelov and X.~Gu,
\newblock ``A cascade of deep learning fuzzy rule-based image classifier and
  svm,''
\newblock in {\em Proc. 2017 IEEE Int. Conf. Systems, Man, and Cybernetics
  (SMC)}. IEEE, 2017, pp. 746--751.

\bibitem{deepTSK}
Y.~Zhang, H.~Ishibuchi, and S.~Wang,
\newblock ``Deep takagi--sugeno--kang fuzzy classifier with shared linguistic
  fuzzy rules,''
\newblock {\em IEEE Trans. Fuzzy Syst.}, vol. 26, no. 3, pp. 1535--1549, 2018.

\bibitem{deepFrcICU}
R.~Davoodi and M.~H. Moradi,
\newblock ``Mortality prediction in intensive care units (icus) using a deep
  rule-based fuzzy classifier,''
\newblock {\em J. Biomedical Informatics}, vol. 79, pp. 48--59, 2018.

\bibitem{dl_honglak_lee}
H.~Lee, P.~Pham, Y.~Largman, and A.~Y. Ng,
\newblock ``Unsupervised feature learning for audio classification using
  convolutional deep belief networks,''
\newblock in {\em Advances in neural information processing systems 21
  (NIPS'09)}, 2009, pp. 1096--1104.

\bibitem{dl_hinton2006}
G.~E. Hinton, S.~Osindero, and Y.~W. Teh,
\newblock ``A fast learning algorithm for deep belief nets,''
\newblock {\em Neural computation}, vol. 18, no. 7, pp. 1527--1554, 2006.

\bibitem{dl_hinton2012}
A.~Krizhevsky, I.~Sutskever, and G.~E. Hinton,
\newblock ``Imagenet classification with deep convolutional neural networks,''
\newblock in {\em Advances in neural information processing systems 24
  (NIPS'12}, 2012, pp. 1097--1105.

\bibitem{dl_fuzzy_rbm}
C.~Chen, C.~Zhang, L.~Chen, and M.~Gan,
\newblock ``Fuzzy restricted boltzmann machine for the enhancement of deep
  learning,''
\newblock {\em IEEE Trans. Fuzzy Syst.}, vol. PP, no. 99, pp. 1--1, 2015.

\bibitem{dl_railway}
O.~Fink, E.~Zio, and U.~Weidmann,
\newblock ``Fuzzy classification with restricted boltzman machines and
  echo-state networks for predicting potential railway door system failures,''
\newblock {\em IEEE Trans. Rel.}, vol. 64, no. 3, pp. 861--868, Sept 2015.

\bibitem{sevakula_sae1}
R.~K. Sevakula and N.~K. Verma,
\newblock ``Fuzzy rule reduction using sparse auto-encoders,''
\newblock in {\em Proc. IEEE Int. Conf. Fuzzy Systems (FUZZ-IEEE'15)}. IEEE,
  2015, pp. 1--6.

\bibitem{sevakula_sae2}
R.~K. Sevakula, A.~Shah, and N.~K. Verma,
\newblock ``Data preprocessing methods for sparse auto-encoder based fuzzy rule
  classifier,''
\newblock in {\em Proc. IEEE Workshop on Computational Intelligence
  (IEEE-WCI'15)}. IEEE, 2015, pp. 1--6.

\bibitem{dl_ng_lecture_notes}
A.~Ng,
\newblock ``Sparse autoencoder,''
\newblock {\em CS294A Lecture notes}, vol. 72, 2011.

\bibitem{dl_denoising_ae}
P.~Vincent, H.~Larochelle, I.~Lajoie, Y.~Bengio, and P.~A. Manzagol,
\newblock ``Stacked denoising autoencoders: Learning useful representations in
  a deep network with a local denoising criterion,''
\newblock {\em The Journal of Machine Learning Research (JMLR)}, vol. 11, pp.
  3371--3408, 2010.

\bibitem{fast_dnn}
S.~Ioffe and C.~Szegedy,
\newblock ``Batch normalization: Accelerating deep network training by reducing
  internal covariate shift,''
\newblock 2015.

\bibitem{categoricalLimitation}
E.~Jang, S.~Gu, and B.~Poole,
\newblock ``Categorical reparameterization with gumbel-softmax,''
\newblock 2016.

\bibitem{learning_MF}
T.~P. Wu and S.~M. Chen,
\newblock ``A new method for constructing membership functions and fuzzy rules
  from training examples,''
\newblock {\em IEEE Transactions on Systems, Man, and Cybernetics, Part B
  (Cybernetics)}, vol. 29, no. 1, pp. 25--40, 1999.

\bibitem{vapnik}
V.~Vapnik,
\newblock {\em Statistical learning theory}, vol.~1,
\newblock Wiley New York, 1998.

\bibitem{cmaes_original}
N.~Hansen, S.~D. M{\"u}ller, and P.~Koumoutsakos,
\newblock ``Reducing the time complexity of the derandomized evolution strategy
  with covariance matrix adaptation (cma-es),''
\newblock {\em Evolutionary computation}, vol. 11, no. 1, pp. 1--18, 2003.

\bibitem{cmaes_implementation}
R.~Ros and N.~Hansen,
\newblock ``A simple modification in cma-es achieving linear time and space
  complexity,''
\newblock in {\em Proc. Int. Conf. Parallel Problem Solving from Nature}.
  Springer, 2008, pp. 296--305.

\bibitem{uci}
M.~Lichman,
\newblock ``Uci machine learning repository,'' 2013.

\bibitem{minfunc_kent}
M.~Schmidt,
\newblock ``minfunc: unconstrained differentiable multivariate optimization in
  matlab,'' 2005.

\bibitem{dl_bengio2012}
Y.~Bengio,
\newblock ``Practical recommendations for gradient-based training of deep
  architectures,''
\newblock in {\em Neural Networks: Tricks of the Trade}, pp. 437--478.
  Springer, 2012.

\bibitem{comparison_quinlan1996}
J.~R. Quinlan,
\newblock ``Improved use of continuous attributes in c4.5,''
\newblock {\em J. Artificial Intelligence Research}, vol. 4, pp. 77--90, 1996.

\bibitem{comparison_elomaa1999}
T.~Elomaa and J.~Rousu,
\newblock ``General and efficient multisplitting of numerical attributes,''
\newblock {\em Machine learning}, vol. 36, no. 3, pp. 201--244, 1999.

\bibitem{comparison_sanchez2001}
L.~S{\'a}nchez, I.~Couso, and J.~A. Corrales,
\newblock ``Combining gp operators with sa search to evolve fuzzy rule based
  classifiers,''
\newblock {\em Information Sciences}, vol. 136, no. 1, pp. 175--191, 2001.

\bibitem{comparison_slave}
A.~Gonzblez and R.~P{\'e}rez,
\newblock ``Slave: A genetic learning system based on an iterative approach,''
\newblock {\em IEEE Trans. Fuzzy Syst.}, vol. 7, no. 2, pp. 176--191, 1999.

\bibitem{slave_overview}
D.~García, A.~González, and R.~Pérez,
\newblock ``Overview of the slave learning algorithm: A review of its evolution
  and prospects,''
\newblock {\em International Journal of Computational Intelligence Systems},
  vol. 7, no. 6, pp. 1194--1221, 2014.

\bibitem{frc_ishibuchi_comparison}
H.~Ishibuchi and T.~Yamamoto,
\newblock ``Comparison of heuristic criteria for fuzzy rule selection in
  classification problems,''
\newblock {\em Fuzzy Optimization and Decision Making}, vol. 3, no. 2, pp.
  119--139, 2004.

\bibitem{comparison_abonyi2003}
J.~Abonyi, J.~A. Roubos, and F.~Szeifert,
\newblock ``Data-driven generation of compact, accurate, and linguistically
  sound fuzzy classifiers based on a decision-tree initialization,''
\newblock {\em Int. J. Approximate Reasoning}, vol. 32, no. 1, pp. 1--21, 2003.

\bibitem{comparison_guan2004}
S.~U. Guan and F.~Zhu,
\newblock ``Class decomposition for ga-based classifier agents-a pitt
  approach,''
\newblock {\em IEEE Trans. Syst., Man, Cybern. Part B}, vol. 34, no. 1, pp.
  381--392, 2004.

\bibitem{frc_ishibuchi_gbml}
H.~Ishibuchi, T.~Yamamoto, and T.~Nakashima,
\newblock ``Hybridization of fuzzy gbml approaches for pattern classification
  problems,''
\newblock {\em IEEE Trans. Syst., Man, Cybern. Part B}, vol. 35, no. 2, pp.
  359--365, 2005.

\bibitem{furia_c}
J.~Hühn and E.~Hüllermeier,
\newblock ``Furia: an algorithm for unordered fuzzy rule induction,''
\newblock {\em Data Mining and Knowledge Discovery}, vol. 19, no. 3, pp.
  293–319, 2009.

\bibitem{wfc_nakashima}
T.~Nakashima, G.~Schaefer, Y.~Yokota, and H.~Ishibuchi,
\newblock ``A weighted fuzzy classifier and its application to image processing
  tasks,''
\newblock {\em Fuzzy Sets and Systems}, vol. 158, no. 3, pp. 284 -- 294, 2007.

\bibitem{FARCHDC_herrera}
J.~Alcala-Fdez, R.~Alcala, and F.~Herrera,
\newblock ``A fuzzy association rule-based classification model for
  high-dimensional problems with genetic rule selection and lateral tuning,''
\newblock {\em IEEE Trans. Fuzzy Syst.}, vol. 19, no. 5, pp. 857--872, 2011.

\bibitem{gfs_adaboost}
M.~J. del Jesus, F.~Hoffmann, L.~J. Navascues, and L.~Sanchez,
\newblock ``Induction of fuzzy-rule-based classifiers with evolutionary
  boosting algorithms,''
\newblock {\em IEEE Trans. Fuzzy Syst.}, vol. 12, no. 3, pp. 296--308, 2004.

\bibitem{gfs_logitboost}
L.~Sánchez and J.~Otero,
\newblock ``Boosting fuzzy rules in classification problems under single-winner
  inference,''
\newblock {\em International Journal of Intelligent Systems}, vol. 22, no. 9,
  pp. 1021--1034, 2007.

\bibitem{ivturs}
J.~A. Sanz, A.~Fernández, H.~Bustince, and F.~Herrera,
\newblock ``Ivturs: A linguistic fuzzy rule-based classification system based
  on a new interval-valued fuzzy reasoning method with tuning and rule
  selection,''
\newblock {\em IEEE Trans. Fuzzy Syst.}, vol. 21, no. 3, pp. 399--411, 2013.

\bibitem{keel}
J.~Alcal{\'a}-Fdez, A.~Fern{\'a}ndez, J.~Luengo, J.~Derrac, S.~Garc{\'\i}a,
  L.~S{\'a}nchez, and F.~Herrera,
\newblock ``Keel data-mining software tool: data set repository, integration of
  algorithms and experimental analysis framework.,''
\newblock {\em J. Multiple-Valued Logic \& Soft Computing}, vol. 17, 2011.

\bibitem{demsar}
Janez Dem{\v{s}}ar,
\newblock ``Statistical comparisons of classifiers over multiple data sets,''
\newblock {\em Journal of Machine learning research}, vol. 7, no. Jan, pp.
  1--30, 2006.

\bibitem{fuzzyhmm}
N.~K. Verma and M.~Hanmandlu,
\newblock ``Additive and nonadditive fuzzy hidden markov models,''
\newblock {\em IEEE Trans. Fuzzy Syst.}, vol. 18, no. 1, pp. 40--56, 2010.

\bibitem{dl_bengio_2007greedy}
Y.~Bengio, P.~Lamblin, D.~Popovici, and H.~Larochelle,
\newblock ``Greedy layer-wise training of deep networks,''
\newblock in {\em Advances in neural information processing systems}, 2007, pp.
  153--160.

\bibitem{dl_lecun_2007}
M.~A. Ranzato, C.~Poultney, S.~Chopra, Y.~L. Cun, et~al.,
\newblock ``Efficient learning of sparse representations with an energy-based
  model,''
\newblock in {\em Advances in neural information processing systems}, 2007, pp.
  1137--1144.

\bibitem{dl_ranzato}
M.~A. Ranzato, Y.~L. Boureau, and Y.~L. Cun,
\newblock ``Sparse feature learning for deep belief networks,''
\newblock in {\em Advances in Neural Information Processing Systems 20
  (NIPS'08)}, pp. 1185--1192. Curran Associates, Inc., 2008.

\end{thebibliography}

\clearpage

\appendices
\section{Autoencoders}

Deep networks have long been known for their expressive power and their ability to understand complex non-linear relationships in data. They are also computationally more efficient than shallow networks. The efficiency comes from the fact that if a $L$ layered Feed Forward Network (FFN) is to be functionally replaced by a single layer FFN, then the number of hidden neurons required in the single layered FFN increases exponentially with $L$ \cite{dl_bengio2009}. Unfortunately even after knowing the power of deep networks, there did not exist enough computation power and methods (except for Recurrent NN) to effectively learn the network weights/bias of the DNN \cite{dl_bengio2009}. Traditional learning of feed forward neural networks 
fails with multiple hidden layers due to: 1) insufficient labeled data, 2) solution getting stuck at poor local optima and 3) long learning time. In 2006, Hinton et al. \cite{dl_hinton2006}, Bengio et al. \cite{dl_bengio_2007greedy} and LeCun et al. \cite{dl_lecun_2007} sparked the deep learning revolution. The advent/use of suitable Graphical Processing Units (GPUs) added to its progress, as it made
related research much quicker to implement. The underlying idea common to all, was to initialize weights with unlabeled data using unsupervised blocks in greedy layer wise manner, followed by fine-tuning of network using labeled data. While RBMs form the unsupervised learning blocks for Deep Belief Networks, AEs form the same for DNNs. Bengio \cite{dl_bengio2009} showed that this kind of greedy layerwise unsupervised learning, not only gives good initialization of weights, but also improves the generalization performance of the entire model. Another significant advantage of this approach is that because learning in the first step is of unsupervised nature, data of similar nature but that not pertaining to the problem at hand, may also be used. This is significantly useful because obtaining labeled data for supervised problem is generally expensive, while unlabeled data of similar nature might be abundantly available. For e.g. the problem of spam email detection needs sample emails to be labeled as spam and not spam, whose size would be much smaller as compared to the entire corpus of unlabeled emails or text.

Autoencoders (AEs) are a class of 3 layered ANNs whose objective is to learn an identity function, where desired output is the input itself. As no class label information is involved during the training, the learning is unsupervised. For reference, sample autoencoders have been shown in Fig. \ref{sae_sample}. Training of AEs begin with initialization of the weight matrix and bias vectors of the network with random or predetermined values. 
The updated weight matrix and bias vector are then stored for future use, so that whenever data needs to be brought to the new representation, it can be easily done through forward propagation till hidden units.

Training of AEs can be seen as a form of non-linear Principal Component Analysis \cite{dl_bengio2009}. When trained under constraints, AEs are known to give useful feature representations at the hidden units. One such constraint is to limit the number of hidden neurons to be fewer than input neurons. This constraint forces the data to be represented by  the fewer hidden neurons, but in a manner where much of the original information is intact for future recovery. Another commonly applied constraint is to enforce sparsity in the data representation. AEs with sparsity constraint are known as Sparse Autoencoders (SAEs). A hidden neuron is considered active when its activation function is significantly greater than 0. Sparsity constraint enforces hidden neurons to not be active most of the times. The purpose of using this constraint is that each training sample be represented by a unique code that is sparse. Such representation makes the hidden units act like discrete variables. Literature \cite{dl_bengio2009,dl_ranzato,dl_bengio2012} shows that sparse over-complete representations are better for some applications, and also are more robust to noise.

Consider an AE with weights $\mathbf{W}$ and biases $\mathbf{b}$. Let $X$ be the input data having $m$ samples and $n$ features, and $\mathbf{O}$ be the target output matrix. Let $L$ be the number of layers in the network and $s_l$ be the number of nodes in $l^{th}$ layer. The cost function for an AE is given by (\ref{NN_cost}). In (\ref{NN_cost}), the first term refers to mean square error between predicted output and desired output, and the second term is for regularization term which is controlled by $\lambda$. The cost function for a SAE is given by (9). Let sparsity parameter $\rho$ be defined as the desired average activation function of hidden units, and let $\widetilde{\rho}$ be the observed average activation function of the hidden neurons. Sparsity constraint as shown by (\ref{kl_div}), uses the KL divergence cost to penalize any deviation of $\widetilde{\rho}$ from $\rho$. The KL divergence penalty is then added with a factor of $\beta$ to (7), to then give the SAE cost function.

\begin{align}
J(\mathbf{W},\mathbf{b}) = &\Big[\frac{1}{m}\sum\limits_{i=1}^{m}\frac{1}{2}\big( \lVert \mathbf{x_i} - \mathbf{o_i}\rVert^2  \big)                    \Big] + \frac{\lambda}{2}\sum\limits_{l=1}^{L-1}\sum\limits_{i=1}^{s_l}\sum\limits_{j=1}^{s_{l+1}}(W_{ji}^{(l)})^2 \nonumber \\ 
& \text{where, } h_{\mathbf{W},\mathbf{b}}(\mathbf{x_i}) = \text{sigmoid}(\mathbf{Wx_i} + \mathbf{b})   \label{NN_cost}
\end{align}
\begin{align}
KL(\rho||\widetilde{\rho}) = \sum\limits_{j=1}^{s_2}\rho log\frac{\rho}{\widetilde{\rho}_j} + (1-\rho) log\frac{(1-\rho)}{(1-{\widetilde{\rho}_j})}
\label{kl_div}
\end{align}
\begin{align}
J_{sparse}(\mathbf{W},\mathbf{b}) = J(\mathbf{W},\mathbf{b}) + \beta\sum\limits_{j=1}^{s_2}KL(\rho||\widetilde{\rho})
\label{sae_cost}
\end{align}

A third constraint available for AEs is the Denoising constraint  \cite{dl_denoising_ae}. It is applied by stochastically corrupting the input data, and desiring the output of AE to be the same as the original, uncorrupted input data. In other words, expected output $\mathbf{O}$ and input data $\mathbf{X}$ are updated in following manner.  
\begin{align}
& \mathbf{O} = \mathbf{X}  \notag \\  
& \mathbf{X} = \mathbf{X} + noise  
\end{align} 
Such a constraint makes the network relatively more robust to noise, and enforces the network to work even when some of the input data is corrupted/missing. AEs trained with this constraint are known as Denoising AEs. It is known from \cite{dl_bengio2009} that RBMs typically have a performance edge over sparse AEs. This is because RBMs can learn manifold representation, while sparse AEs cannot. Vincent et al. \cite{dl_denoising_ae} showed that the inclusion of denoising constraint, allows AEs to learn manifold representation, and therefore perform comparable to RBMs.

Larochelle et. al. \cite{dl_ng_lecture_notes} presented stacked AEs, where AEs are learned in layerwise fashion to pre-train a DNN (or initialize the weights and biases of the DNN). Before pre-training, the network architecture, i.e. the number of hidden layers, and the number of nodes in each layer is decided. 
Stacked AEs are then built in the following manner. An AE is first built over the input layer and the first hidden layer of the DNN. The AE is then trained upon the unlabelled data, to learn the initial weights and biases between the input layer and the first hidden layer. All the unlabeled training data is then forward propagated from the input layer to the first hidden layer, and the data is accordingly transformed to a new feature representation. The transformed data is now used to train a second AE. The second AE has the first hidden layer as its input layer, and the second hidden layer of the DNN as its hidden layer. The second AE is then trained upon the transformed data, to learn the initial weights and biases between the first hidden layer and the second hidden layer. The unlabelled data can now be forward propagated till the second hidden layer, to obtain a second level feature representation. 
The process of learning deeper networks, and creating higher level feature representation in greedy layerwise fashion, may continue further as per the DNN architecture. This process of layerwise learning of AEs can be graphically seen in Fig.\ref{sae_sample}. In the case when denoising sparse AEs are used as the building blocks, the stacked network is called Stacked Denoising Sparse Autoencoders (SDSAE).

\section{FRC used on the generated features} 
The procedure for training an FRC following Ishibuchi et al. \cite{frc_ishibuchi2005} on a multi-class problem having $P$ class labels is briefly described in Algorithm 1. The notations are as follows. $\mathbf{X}$ shall represent the data being learned with $m$ data samples and $n$ features. $i$ would be used to represent a sample number, $j$ would represent a feature, $p$ would represent a class label, and $h$ would represent a fuzzy rule.

\noindent\rule{8.8cm}{0.4pt} \\
Algorithm 1 : Generating Fuzzy Rules for Classification of Data \cite{frc_ishibuchi2005}\\
\noindent\rule{8.8cm}{0.4pt}

\begin{description}
	
	\item [\textit{Step 1}:]~ Each dimension of the input vector is divided into as many membership functions as needed. The membership functions until specified are assumed to be Gaussian MFs. The consequent class label can be any one of the $P$ labels.  \\[-1.5ex]
	
	\item [\textit{Step 2}:]~ Given a data sample, the fuzzy regions where the input vector obtain maximum MV, those regions are recognized and noted down. A fuzzy rule is then accordingly formulated, where antecedent part of the rule is formed with the recognized fuzzy regions of previous step, and the consequent class of the rule is decided to be the same as that of the data sample's class label. In this manner, fuzzy rules for all training data are formulated. 
	\\[-1.5ex]
	
	\item [\textit{Step 3}:]~ From the generated rules, the unique rules are found. If there exists more than one rule (which have different consequents) for a given antecedent, then calculate $\beta_{p}$ for each consequent $p$, 
	\begin{equation}
	\beta_{p} = \sum_{y_i\in p}\prod_{j=1}^{n}\mu_{ij}(x_{ij})       
	\end{equation}
	where $y_i\in p$ refer to all data samples giving this rule with $p$ as the output class label, $M$ refers to the number of consequent classes that face conflict with this rule, and $\mu_{ij}()$ is the MF of the region that $x_{ij}$ is most closely associated with.    \\[-1.5ex]
	
	\item [\textit{Step 4}:]~ For resolving conflicts amongst rules, the class label having highest $\beta_{p}$ value is decided to be the consequent class label $p'$ of the rule. \\[-1.5ex]
	
	\item [\textit{Step 5}:]~ Grade of certainties are assigned to each of the final rules. Grade of Certainty $CF_{h}$ of $h^{\text{th}}$ rule with $p'$ as consequent class label, is given by (when $P\geq3$)
	\begin{equation}
	\beta = \sum_{p\not= p'}\frac{\beta_{p}}{M-1} ~~ ,~~
	CF_{h}  =\frac{\beta_{p'}-\beta}{\sum\limits_{p=1}^{M}\beta_{p}}
	\end{equation}

\end{description}
\vspace{-1.0ex}
\noindent\rule{8.8cm}{0.4pt} \\ [-1.5ex]

Test Phase - For finding the class label of a test input vector $\mathbf{x_i}$, calculate $\alpha_{P}$ for each class using (\ref{frc_alpha}). The class receiving maximum $\alpha$, is selected as the predicted class of the sample. 
\begin{align}
&\alpha_{P} = \sum_{{h}\in P} \Bigg( \sum\limits_{j=1}^{n} \mu_{hj} (x_{ij})  \Bigg) .CF_{h}  
\label{frc_alpha}
\end{align}


\section{Case Study - Using Expert Knowledge}
\begin{figure}[htbp]
	\centering
	\includegraphics[scale = 0.25]{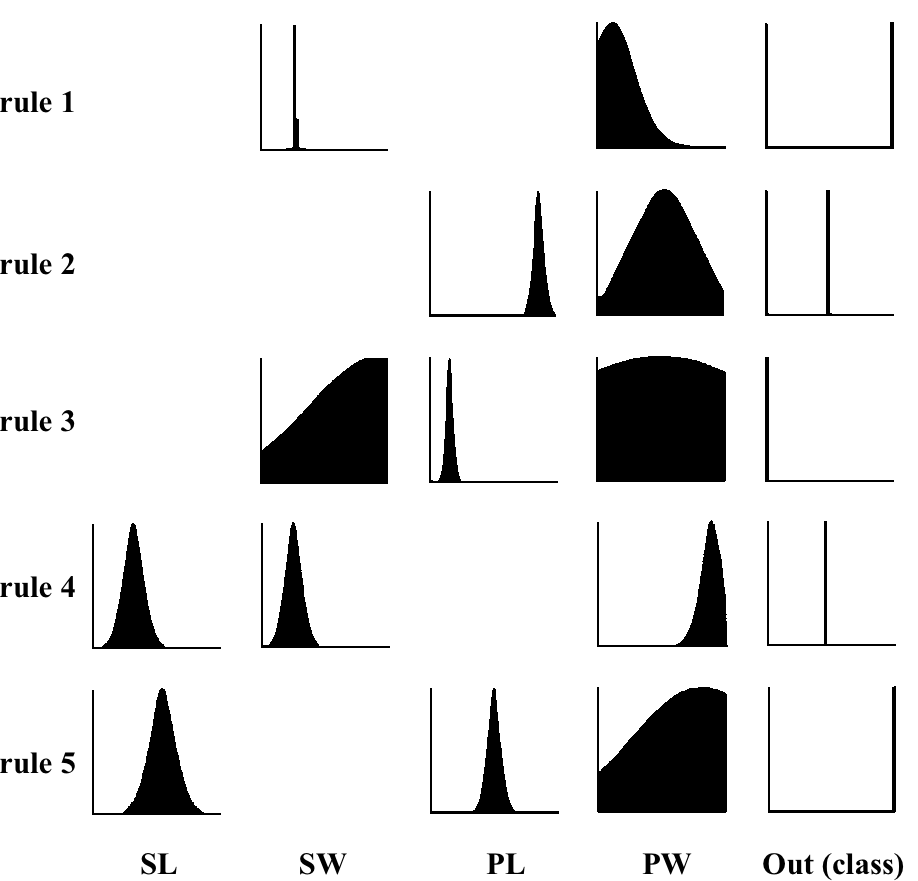}
	\vspace{-1em}
	\caption{Expert rules mentioned in \cite{neuro_fugenesys} for classification of Iris data. Features 1-4 i.e. sepal length(SL), sepal width(SW), petal length(PL) and petal width(PW) vary in the range (in cm) of [4.3,7.9], [2.0,4.4], [1.0,6.9] and [0.1,2.5] respectively.}
	\vspace{-0.5em}
	\label{expert_plots}
\end{figure}

The figure showing the expert rules from \cite{neuro_fugenesys} is placed in Fig. \ref{expert_plots} next page. On interpreting the figure, multiple Gaussian expert MFs were retrieved from each feature for later use as preprocessing functions in proposed framework. The MF parameters that were retrieved from the rules are mentioned in Table \ref{expert_mfs}. Further, following the 5 rules, 5 expert samples were also retrieved. These expert samples are shown in Table \ref{expert_samples}. As expected their form is that of sparse binary vectors.


\begin{table}[htbp]  
	
	\centering
	\caption{Expert gaussian MF parameters retrieved from Fig. \ref{expert_plots}}
	\vspace{-0.5em}
	\begin{tabular}{c|c|c|c|c} 
		\hline \hline
		Expert MF & Feature & \# MF & $\mu$ (in cm) & $\sigma$ (in cm) \\
		
		\hline
		
		$\text{MF}_{11}$ & SL & 1 & 5.429 & 0.706 \\  
		$\text{MF}_{12}$ & SL & 2 & 6.347 & 0.882 \\
		$\text{MF}_{21}$ & SW & 1 & 2.682 & 0.0165 \\
		$\text{MF}_{22}$ & SW & 2 & 4.235 & 2.118 \\
		$\text{MF}_{23}$ & SW & 3 & 2.612 &	0.329 \\
		$\text{MF}_{31}$ & PL & 1 & 6.148 &	0.463 \\
		$\text{MF}_{32}$ & PL & 2 & 1.866 &	0.289  \\
		$\text{MF}_{33}$ & PL & 3 & 3.950 & 0.550 \\
		$\text{MF}_{41}$ & PW & 1 & 0.359 & 0.729 \\
		$\text{MF}_{42}$ & PW & 2 & 1.394 &	1.176 \\
		$\text{MF}_{43}$ & PW & 3 & 1.276 &	4.706 \\
		$\text{MF}_{44}$ & PW & 4 & 2.288 & 0.424  \\
		$\text{MF}_{45}$ & PW & 5 & 2.076 & 2.588  \\
		
		\hline \hline
	\end{tabular}
	\vspace{-0.5em}
	\label{expert_mfs}
\end{table}



\vspace{-5em}


\begin{table*}[ht]  
	\scriptsize
	\centering
	\caption{Expert samples for Iris data}
	\vspace{-1em}
	\begin{tabular}{c |
			*{1}{>{\centering\arraybackslash\rule{0pt}{0.50em}}m{0.6cm}} |
			*{1}{>{\centering\arraybackslash\rule{0pt}{0.50em}}m{0.5cm}} |
			*{1}{>{\centering\arraybackslash\rule{0pt}{0.50em}}m{0.5cm}} |
			*{1}{>{\centering\arraybackslash\rule{0pt}{0.50em}}m{0.5cm}} |
			*{1}{>{\centering\arraybackslash\rule{0pt}{0.50em}}m{0.5cm}} |
			*{1}{>{\centering\arraybackslash\rule{0pt}{0.50em}}m{0.5cm}} |
			*{1}{>{\centering\arraybackslash\rule{0pt}{0.50em}}m{0.5cm}} |
			*{1}{>{\centering\arraybackslash\rule{0pt}{0.50em}}m{0.5cm}} |
			*{1}{>{\centering\arraybackslash\rule{0pt}{0.50em}}m{0.5cm}} |
			*{1}{>{\centering\arraybackslash\rule{0pt}{0.50em}}m{0.5cm}} |
			*{1}{>{\centering\arraybackslash\rule{0pt}{0.50em}}m{0.5cm}} |
			*{1}{>{\centering\arraybackslash\rule{0pt}{0.50em}}m{0.5cm}} |
			*{1}{>{\centering\arraybackslash\rule{0pt}{0.50em}}m{0.5cm}} | c}
		
		\hline \hline
		
		\# Sample & $\text{MF}_{11}$ & $\text{MF}_{12}$ & $\text{MF}_{21}$ & $\text{MF}_{22}$ & $\text{MF}_{23}$ & $\text{MF}_{31}$ & $\text{MF}_{32}$ & $\text{MF}_{33}$ & $\text{MF}_{41}$ & $\text{MF}_{42}$ & $\text{MF}_{43}$ & $\text{MF}_{44}$ & $\text{MF}_{45}$ & Output Class \\
		\hline 
		
		1 & 0 & 0 & 1 & 0 & 0 & 0 & 0 & 0 & 1 & 0 & 0 & 0 & 0 & Versicolor \\ 
		
		2 & 0 & 0 & 0 & 0 & 0 & 1 & 0 & 0 & 0 & 1 & 0 & 0 & 0 & Virginica \\ 
		
		3 & 0 & 0 & 0 & 1 & 0 & 0 & 1 & 0 & 0 & 0 & 1 & 0 & 0 & Setosa \\ 
		
		4 & 1 & 0 & 0 & 0 & 1 & 0 & 0 & 0 & 0 & 0 & 0 & 1 & 0 & Virginica \\ 
		
		5 & 0 & 1 & 0 & 0 & 0 & 0 & 0 & 1 & 0 & 0 & 0 & 0 & 1 & Versicolor \\ 
		
		\hline \hline
	\end{tabular}
	\label{expert_samples}
	\vspace{-1em}
\end{table*}


\vspace {3cm}
\section{Experiments on Additional Datasets}

To establish the effectiveness of the proposed approach, in this section, we conduct more experiments and then perform statistical tests. Table \ref{datasets_more} provides details of the 10 datasets, on whom the additional experiments were performed. 
All datasets are available at the KEEL dataset repository \cite{keel}. We found our approach to specifically work well when the number of classes were 2 or 3; hence most of our datasets are 2-classed or 3-classed. For datasets which had more than 3 classes, the positive class and negative class were selected, and their details are mentioned in the "[PosClass, NegClass]" column.

\vspace{-3cm}

\begin{table}[t]
	\centering
	\caption{Details of Experimented Datasets}
	\vspace{-0.5cm}
	\begin{tabular}{*{1}{>{\centering\arraybackslash\rule{0pt}{0.50em}}m{1.5cm}} 
			*{1}{>{\centering\arraybackslash\rule{0pt}{0.50em}}m{0.9cm}} 
			*{1}{>{\centering\arraybackslash\rule{0pt}{0.50em}}m{0.9cm}} 
			*{1}{>{\centering\arraybackslash\rule{0pt}{0.50em}}m{0.8cm}}  
            *{1}{>{\centering\arraybackslash\rule{0pt}{0.50em}}m{2.5cm}}}\\                         
		\hline \hline
		Datasets & Attributes & Samples & Classes & [PosClass, NegClass] \\
		\hline
		Appendicitis & 7 & 106 & 2 & - \\
		Australian & 14 & 690 & 2 & - \\
		Balance & 4 & 625 & 3 & - \\
		Letter & 17 & 1555 & 2 & ["A","B"] \\
		Segment & 19 & 660 & 2 & ["Cement","Window"] \\
		Movement\-Libras & 90 & 48 & 2 & [1,2] \\
		Optdigits & 64 & 1128 & 2 & [1,2] \\
		Spambase  & 57 & 4597 & 2 & - \\
		Spectfheart & 44 & 267 & 2 & - \\
		WDBC & 30 & 569 & 2 & - \\
		\hline \hline
	\end{tabular}
	\label{datasets_more}
	\vspace{-0.5em}
\end{table}


\begin{table}[t]
	\centering
	\vspace{-0.4cm}
	\caption[caption]{Detailed Results with Proposed Methods (in \%)}
	\vspace{-0.5cm}
	\begin{tabular}{*{1}{>{\centering\arraybackslash\rule{0pt}{0.50em}}m{1.5cm}} | c |
			*{1}{>{\centering\arraybackslash\rule{0pt}{0.50em}}m{1.7cm}} 
            *{1}{>{\centering\arraybackslash\rule{0pt}{0.50em}}m{1.7cm}}}\\		
		\hline \hline
		
		Dataset & Param & FT-I & FT-IV \\ [0.5ex]

		\hline
		
		
		\multirow{3}{*}{Appendicitis}  & Accuracy & 87.82\rpm6.29  &  88.64\rpm7.83 \\[0.5ex]

		&  Rule Size , $\rho$ & 8.8, 0.9 & 3,0.6 \\[-0.3ex]
		
		& Net. Arch.  & (7,5) & (7,5) \\[1.5ex] 
		
		
		\multirow{3}{*}{Australian}  & Accuracy & 86.09\rpm4.39  &  86.52\rpm4.43\\[0.5ex]

		&  Rule Size , $\rho$ & 4.5,0.5 & 10.5,0.1 \\[-0.3ex]
		
		& Net. Arch.  & (14,11)  & (14,11) \\[1.5ex] 
		
		
		\multirow{3}{*}{Balance}  & Accuracy & 91.67\rpm3.55  &  91.35\rpm3.89\\[0.5ex]

		&  Rule Size , $\rho$ & 3,0.2 & 3.76,0.4 \\[-0.3ex]
		
		& Net. Arch.  & (4,4)  & (4,4) \\[1.5ex]

		
		\multirow{3}{*}{Letter}  & Accuracy & 99.45\rpm0.47  &  99.16\rpm0.53\\[0.5ex]

		&  Rule Size , $\rho$ & 2.1,0.3 & 11,0.5 \\[-0.3ex]
		
		& Net. Arch.  & (17,14,11) & (17,14) \\[1.5ex]

		
		\multirow{3}{*}{Segment}  & Accuracy & 96.21\rpm3.05  &  95.76\rpm2.56\\[0.5ex]

		&  Rule Size , $\rho$ & 2.3,0.7 & 3.3,0.8 \\[-0.3ex]
		
		& Net. Arch.  & (19,15,11) & (19,15,11) \\[1.5ex] 
		
		
		\multirow{3}{*}{Libras}  & Accuracy & 83.00\rpm17.67  &  85.50\rpm10.12\\[0.5ex]

		&  Rule Size , $\rho$ & 18.1,0.3 & 7.8,0.2 \\[-0.3ex]
		
		& Net. Arch.  & (90,40)  & (90,40) \\[1.5ex] 
		
		
		\multirow{3}{*}{Optdigits}  & Accuracy & 98.05\rpm2.20  &  98.67\rpm0.75\\[0.5ex]

		&  Rule Size , $\rho$ & 3.7,0.5 & 26.6,0.3 \\[-0.3ex]
		
		& Net. Arch.  & (64,50)  & (64,50) \\[1.5ex] 
		
		
		\multirow{3}{*}{Spambase}  & Accuracy & 83.16\rpm3.45  &  86.27\rpm4.32\\[0.5ex]

		&  Rule Size , $\rho$ & 2193.4,0.8 &  1593.5,0.5 \\[-0.3ex]
		
		& Net. Arch.  & (57,40)  & (57,40) \\[1.5ex] 
		
		
		\multirow{3}{*}{Spectfheart}  & Accuracy & 77.14\rpm7.24  &  79.76\rpm7.55\\[0.5ex]

		&  Rule Size , $\rho$ & 132.8,0.2 &  136.2,0.3 \\[-0.3ex]
		
		& Net. Arch.  & (44,30)  &  (44,30) \\[1.5ex] 
		
		
		\multirow{3}{*}{WDBC}  & Accuracy & 96.84\rpm3.07  &  97.37\rpm2.89\\[0.5ex]

		&  Rule Size , $\rho$ & 5.3,0.5 &  9,0.3 \\[-0.3ex]
		
		& Net. Arch.  & (30,22)  &  (30,22) \\[1.5ex] 
		

		\hline \hline
	\end{tabular}
	\label{results_comp_more}
    \vspace{-1.5em}
\end{table}



\begin{table*}
	\centering
	\caption[caption]{Statistical Analysis with Test classification error }
	\begin{tabular}{c|c c c c c | c c c c c} 
		
		\hline \hline
		
		\multirow{2}{*}{Dataset} & \multicolumn{5}{c|}{Test Classification Error (in \%)} & \multicolumn{5}{c}{Ranks}\\ [0.8ex]
		
		& 
		
		\begin{tabular}[c]{@{}c@{}}Slave\_v0\end{tabular} &  
		\begin{tabular}[c]{@{}c@{}}SGERD\end{tabular} & 
		\begin{tabular}[c]{@{}c@{}}ChiRW\end{tabular} & 
		\begin{tabular}[c]{@{}c@{}}IVTURS\end{tabular} & 
		\begin{tabular}[c]{@{}c@{}}Proposed\end{tabular} &  
		\begin{tabular}[c]{@{}c@{}}Slave\_v0\end{tabular} & 
		\begin{tabular}[c]{@{}c@{}}SGERD\end{tabular} & 
		\begin{tabular}[c]{@{}c@{}}ChiRW\end{tabular} &
		\begin{tabular}[c]{@{}c@{}}IVTURS\end{tabular} & 
		\begin{tabular}[c]{@{}c@{}}Proposed\end{tabular} \\
		
		\hline
		Iris    & 4.00  &  3.07   &  7.34  &  3.33  &  2.00     &  4  &  2  &  5  &  3  & 1  \\ [0.5ex] 
		Wine    & 6.18  &  3.81   & 6.18  & 7.39   &  3.4    &  3.5    & 2     &  3.5   & 5    & 1     \\ [0.5ex] 
		Cancer  &  6.30   & 2.98   & 7.92     & 3.22   & 2.2   & 4    & 2     & 5   & 3   & 1   \\ [0.5ex] 
		Sonar  & 30.26  & 22.80   & 39.95  & 21.71  & 16.38  & 4   & 3  & 5   & 2   & 1      \\ [0.5ex] 
		Pima & 23.82   & 25.36   & 26.93   & 25.63  & 21.62   & 2   & 3    & 5   & 4    & 1     \\ [0.5ex] 
		Appendicitis  & 18.00  & 13.09   &  14.19  & 15.00  & 11.36   &  5  & 2   & 3   & 4   & 1     \\ [0.5ex] 
		Australian   & 15.65  & 14.49  & 20.14   & 13.53  & 13.48   & 4   & 3   & 5   & 2   &  1  \\ [0.5ex] 
		Balance  & 31.86  & 24.02   & 9.44   & 12.31  & 8.33   &  5  & 4    & 2   & 3   & 1       \\ [0.5ex] 
		Letter  & 2.25  & 5.66    &  1.55   & 1.48  &  0.57   & 4   & 5   & 3   & 2   & 1      \\ [0.5ex] 
		Segment  & 10.00  & 9.85  & 9.55  & NA  & 3.79   &  4   & 3   & 2   &  NA   & 1      \\ [0.5ex] 
		Libras &  35.00   & 24.00  & 22.50  & 6.00  &  14.50   & 5  & 4   & 3   &  1  &  2 \\ [0.5ex]
		Optdigits  & 2.39  & 6.02  & 5.07  &  2.06   &  1.33   & 3  & 5   & 4   &  2  &  1   \\ [0.5ex] 
		Spambase  &  21.62   & 29.61  & 28.54 & 10.18 &  13.73 & 3  & 5  & 4    & 1   &  2     \\ [0.5ex] 		
		Spectfheart  & 22.08  &  21.33 & 34.10  &  20.47   & 20.24    &  4 & 3   & 5    & 2   &  1  \\ [0.5ex] 		
		WDBC  &  4.752   & 8.61  & 7.03  &   4.40  &  2.63   & 3  & 5   & 4   & 2   &  1    \\ [0.5ex] 
		Sum of Ranks &&&&&& 57.5 & 51 & 58.5 & 36 & 17\\ [0.5ex] 
		Average Rank &&&&&& 3.83 & 3.4 & 3.9 & 2.57 & 1.13\\ [0.5ex] 
		\hline \hline
	\end{tabular}
	\label{statisticalAnalysis}
	
\end{table*}

\vspace{3cm}

The results achieved with proposed approach are presented in Table \ref{results_comp_more}. As mentioned earlier, due to computational requirements, only the FT-I and FT-IV strategies were evaluated here. For statistical comparisons, we chose four recent and popular FRCs, namely Slave\_v0 \cite{slave_overview}, SGERD \cite{reduction_sgerd}, ChiRW\cite{frc_ishibuchi2005}, and IVTURS \cite{ivturs}. Their average test classification errors are presented in Table \ref{statisticalAnalysis}. The implementations present in the KEEL software \cite{keel} were used to generate these results. In the table, the best among the results achieved with FT-I and FT-IV strategies, are shown under the "Proposed" column for future comparison. It may be noted that the test performance of each approach was evaluated over 10-fold cross-validation. The cross validation sets used in our experiments are made available online at ``http://iitk.ac.in/idea/datasets/sdsaefrc".

We were unable to ascertain the performance of IVTURS on Segment data. On the Segment data, IVTURS provided irrelevant results (Null values). 

Statistical analyses/tests are commonly used for comparing the performance of multiple approaches over multiple datasets. Demsar et al.'s paper \cite{demsar} is a widely cited work that describes popular statistical tests for comparing classifiers. Here, we use the Friedman test, the Bonferroni-Dunn posthoc test, and the Wilcoxon Signed Ranks test to infer conclusions. Demsar et al. \cite{demsar} state that using the performance values as they are for statistical comparisons, can create doubts on the commensurability. A reasonable approach therefore is to rank the classifier performances of each approach, for each dataset, and then use those ranks for statistical analyses. Table \ref{statisticalAnalysis}, along with test classification errors, presents the respective ranks of each approach for each dataset. Approaches with lower test classification error are ranked higher, with "1" being the highest rank and "5" being the lowest rank. The ranks for each approach were summed and their average rank was found. While computing the average ranks for an approach, the datasets where the approach could not provide a result were omitted, e.g. average rank for IVTURS was 36/14 = 2.57.

Performing Friedman test with the average ranks $\mathbf{R}$, $N$ datasets, and $l$ classifier comparisons, 
\begin{align}
\chi_F^2 &= \frac{12N}{l(l+1)}\Bigg[\sum\limits_{j=1}^{l}R_j^2 - \frac{l(l+1)^2}{4}\Bigg]  \label{chi_sq} \nonumber  \\[0.5em]
 & =  \frac{12\times15}{5\times6}\Bigg[3.83^2 + 3.4^2 + 3.9^2 + 2.57^2 + 1.13^2 - \frac{5\times6^2}{4}\Bigg] \nonumber \\[0.5em]
& = ~ 25.9242  \nonumber 
\end{align}
The degrees of freedom would be $l-1$ i.e. 4. Looking at the $\chi^2$ distribution values, the cut-off required to reject Null Hypothesis at $\alpha=0.05$ is 9.488. Since our $\chi_F^2$ is greater than the cut-off, the Null Hypothesis can be rejected. In fact, Null Hypothesis can even be rejected at $\alpha=0.001$, whose required cut-off value is 18.467. A Null Hypothesis is a state, where the differences in average ranks are considered to be purely random, and that there is no statistical significance for the given value of $\alpha$. Rejecting it, states that the difference have strong statistical significance. 

Now that the Null Hypothesis is rejected, we perform the Bonferroni-Dunn posthoc test to specifically compare the proposed approach with the other four approaches. The Bonferroni-Dunn test allows one to compute a required Critical Difference (CD) value for a given $\alpha$. For the proposed approach to be statistically better than another, its average rank must be lower than others', by the CD value. The CD for our case, with $q_{0.05}$ for 4 degrees of freedom being 2.394, is
\begin{align}
CD & = q_\alpha\sqrt{\frac{l(l+1)}{6N}} \nonumber \\
& = 2.394 \times \sqrt{\frac{5\times6}{6\times15}}  ~~ = 1.382 \nonumber  
\end{align}
Computing the difference in average ranks of proposed approach with others, we have the difference to be 1.44 with IVTURS, 2.77 with ChiRW, 2.27 with SGERD, and 2.70 with Slave\_v0. Hence it can be concluded that the proposed approach performs statistically better than ChiRW, SGERD, IVTURS, and Slave\_v0 at $\alpha=0.05$. 

The Wilcoxon Signed-Ranks test for pairwise comparison is one of the easiest statistical tests. To statistically compare a pair of approaches, we count the wins and losses by each approach. Given the number of datasets, $N$, the cut-off for minimum number of wins to be statistically better, are made available at various values of $\alpha$. Few cut-off values are given in Table 3 of \cite{demsar}. For our test here, we can count the wins and losses of the proposed approach with IVTURS on 14 datasets, with ChiRW on 15 datasets, with SGERD on 15 datasets, and with Slave\_v0 on 15 datasets. The proposed approach wins on 12 datasets with IVTURS, 15 datasets with ChiRW, 15 datasets with SGERD, and 15 datasets with Slave\_v0. The cut-off values for minimum wins at $\alpha=0.05$ for 14 datasets and 15 datasets are 11 and 12 respectively. Therefore, it can be easily ascertained that the proposed approach performs statistically better than IVTURS, ChiRW, SGERD and Slave\_v0 at $\alpha=0.05$.

\end{document}